\pdfoutput=1

\documentclass[journal]{IEEEtran}
\usepackage{graphicx}
\usepackage{amsmath}
\usepackage{amssymb}
\usepackage{booktabs}
\usepackage{multirow}
\newcommand{\xmark}{\ding{55}} 
\usepackage{pifont}
\usepackage{pifont}
\newcommand{\cmark}{\ding{51}}  
\usepackage{algorithm}
\usepackage{algpseudocode}

\ifCLASSINFOpdf
\else
\fi
\begin{document}

%
\title{RAH-VLA: Resolution-Adaptive Hierarchical Vision–Language Alignment
for Multimodal Remote Sensing Understanding}
%
%
%


\author{Siyu Zhang, Lianlei Shan, Runhe Qiu, and Wenxin Zhong
\thanks{This work was supported by the Sanda University Special Research Fund
Project under Grant 2024BSZX05. (Corresponding author: Wenxin Zhong.)}
\thanks{S. Zhang, R. Qiu, and W. Zhong are with Sanda University,
Shanghai 201209, China
(e-mail: zhangsiyu0066@gmail.com; qiurh@sandau.edu.cn;
wxzhong@sandau.edu.cn).}
\thanks{L. Shan is with Tsinghua University, Beijing 100084, China
(e-mail: shanlianlei18@mails.ucas.edu.cn).}
}

%
%

\markboth{}%
{Zhang \MakeLowercase{\textit{et al.}}: RAH-VLA: Resolution-Adaptive Hierarchical Vision–Language Alignment for Multimodal Remote Sensing Understanding}
%



\maketitle


\begin{abstract}
Multimodal vision--language modeling has emerged as a promising paradigm for remote sensing (RS) image understanding. However, existing methods are limited by fixed-resolution visual processing and single-scale vision--language alignment, making it difficult to simultaneously preserve fine-grained details and maintain semantic consistency across different spatial granularities. To address these challenges, we propose RAH-VLA, a Resolution-Adaptive Hierarchical Vision--Language Alignment framework for multimodal remote sensing understanding. Specifically, a Dynamic Resolution Input Strategy (DRIS) is developed to enable resolution-adaptive visual representations, while a Multi-scale Vision--Language Alignment Mechanism (MS-VLAM) is introduced to establish hierarchical semantic correspondence across object-level, region-level, and global-level representations. Extensive experiments on multiple remote sensing benchmarks demonstrate that RAH-VLA consistently improves image captioning, visual grounding, and cross-modal reasoning performance while reducing computational redundancy. Qualitative analyses further illustrate the effectiveness of the proposed resolution-adaptive perception and hierarchical vision--language alignment mechanisms, as well as the cross-modality generalization capability of the proposed framework. Overall, RAH-VLA provides an effective and scalable solution for multimodal remote sensing interpretation, offering a practical pathway toward efficient and semantically robust RS vision--language models.
\end{abstract}

\begin{IEEEkeywords}
multimodal remote sensing, vision-language modeling, hierarchical alignment, resolution adaptation, cross-modal semantic understanding
\end{IEEEkeywords}

%
\IEEEpeerreviewmaketitle

\section{Introduction}

\begin{figure*}[!t]
\centering
\includegraphics[width=0.9\textwidth]{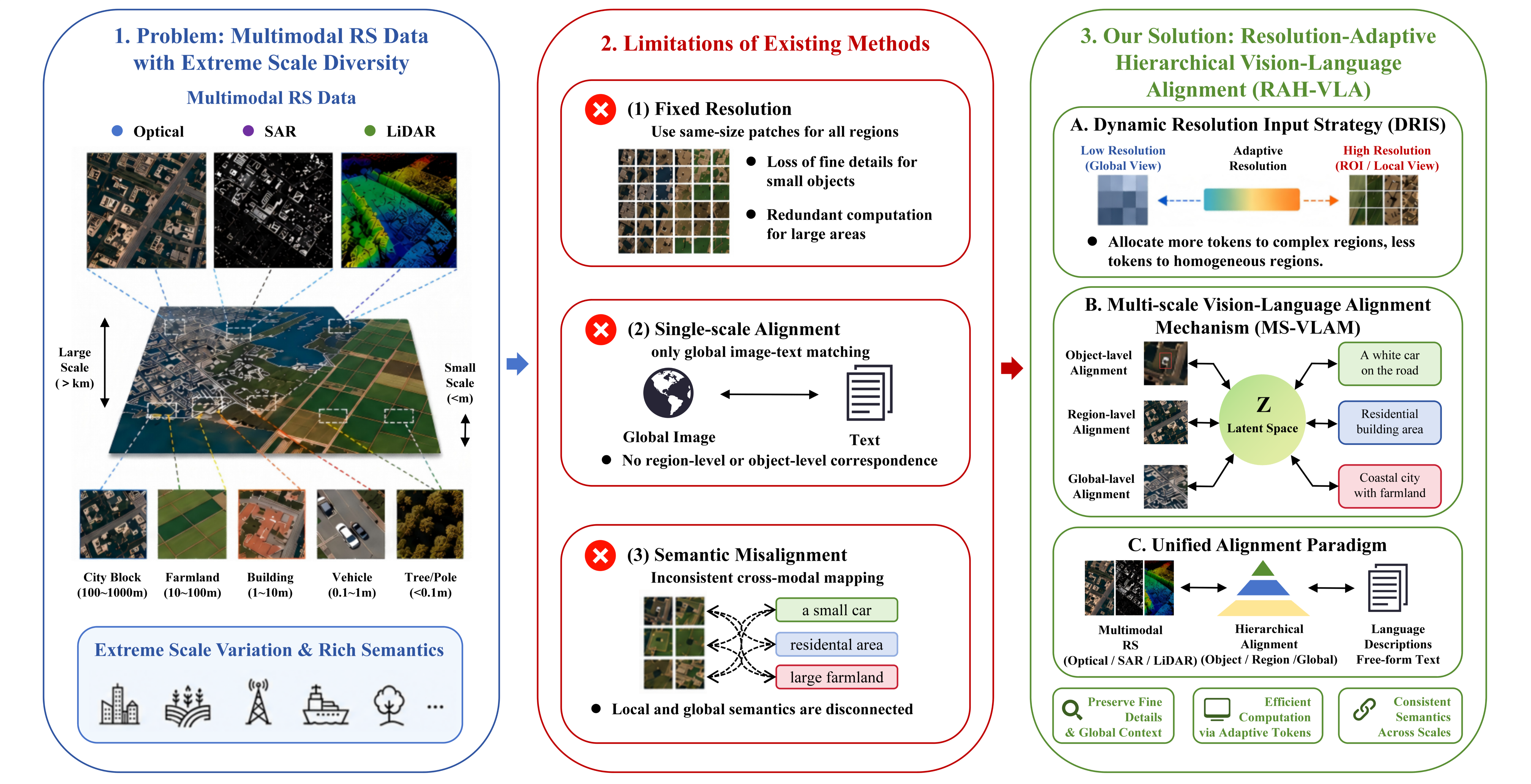}
\caption{
\textbf{Motivation and overview of the proposed RAH-VLA framework for remote sensing image understanding.}
Remote sensing imagery exhibits heterogeneous modalities, extreme scale variation, and hierarchically distributed semantics, posing challenges for multimodal understanding. Existing methods are limited by fixed-resolution processing and single-scale alignment. RAH-VLA provides a unified framework for jointly modeling resolution-adaptive visual perception and hierarchical vision--language semantic alignment.
}
\label{fig_1}
\end{figure*}

\IEEEPARstart{R}{emote} sensing (RS) imagery, acquired from satellite-borne and airborne platforms, provides a fundamental means for large-scale Earth observation and dynamic environment monitoring \cite{ref1_zs}. Benefiting from wide spatial coverage, high spatiotemporal resolution, and rich spectral diversity, modern RS data supports a broad range of applications, including land-cover analysis, urban management, precision agriculture, and disaster monitoring \cite{ref62,ref2_app1,ref3_app2}. With the rapid advancement of sensing technologies and Earth observation systems, remote sensing imagery is evolving toward increasingly large-scale and heterogeneous data distributions, driving growing demands for intelligent semantic understanding \cite{ref63}.

Despite the rapid advancement of sensing technologies and Earth observation systems, achieving reliable semantic understanding of remote sensing imagery remains highly challenging, particularly in scenarios requiring the joint perception of large-scale spatial context and fine-grained object details \cite{ref4,ref5}. Traditional unimodal analysis is inherently insufficient to capture the complementary and heterogeneous information distributed across modern RS data (e.g., optical, SAR, and LiDAR modalities), motivating the transition toward multimodal semantic understanding paradigms \cite{ref7,ref8}. Existing multimodal approaches mainly rely on implicit feature interaction and feature-level fusion, making it difficult to establish consistent semantic correspondence across heterogeneous modalities and multiple semantic granularities \cite{ref13,ref64}. Consequently, these methods often struggle to achieve robust high-level semantic understanding in complex remote sensing scenarios.

To overcome the limitations of conventional multimodal fusion, vision-language models (VLMs) have recently emerged as a promising paradigm for cross-modal semantic understanding by explicitly bridging visual representations and language-based semantic concepts \cite{ref17,ref18}. Compared with traditional multimodal frameworks, VLMs provide a unified framework for connecting visual content with semantic concepts, enabling high-level tasks such as image captioning, visual grounding, and multimodal reasoning \cite{ref19,ref20}. More importantly, VLMs enable explicit modeling of semantic relationships beyond low-level feature interaction, offering a principled way to align visual representations with language-guided semantics \cite{ref21,ref22}.

Despite their success, existing VLMs are primarily designed for natural image domains and do not explicitly account for the intrinsic characteristics of remote sensing imagery \cite{ref5,ref1_zs}. In particular, RS imagery exhibits extreme scale variation, where large homogeneous regions coexist with small but semantically critical objects, making fixed-resolution processing inefficient and prone to losing fine-grained semantic details \cite{ref25}. Meanwhile, semantic concepts in RS imagery are inherently spatially distributed and hierarchically structured across multiple scales \cite{ref15}. However, most existing VLMs mainly rely on single-scale alignment, limiting their ability to capture consistent cross-modal semantics across object-level, region-level, and global-level representations \cite{ref65}. In addition, the weak and indirect correspondence between visual patterns and textual descriptions further increases the difficulty of robust cross-modal semantic alignment under heterogeneous sensing conditions \cite{ref19}.

These limitations reveal a fundamental mismatch between resolution-dependent visual representations and hierarchically structured semantic concepts in remote sensing imagery, where fixed-resolution visual encoding fails to preserve consistent semantic correspondence across spatial scales and semantic granularities \cite{ref26}. As illustrated in Fig.~\ref{fig_1}, multimodal remote sensing imagery exhibits extreme scale variation and hierarchically distributed semantic structures, while existing vision--language frameworks mainly rely on fixed-resolution processing and single-scale alignment, resulting in inefficient representation learning and inconsistent semantic correspondence across different spatial granularities. These observations highlight the necessity of a unified modeling paradigm that can simultaneously account for scale variation and hierarchical semantic structures in remote sensing \cite{ref27}.

To address these challenges, we formulate multimodal remote sensing understanding as a \emph{resolution-adaptive hierarchical vision--language alignment} problem and propose \textbf{RAH-VLA} (\textbf{R}esolution-\textbf{A}daptive \textbf{H}ierarchical \textbf{V}ision--\textbf{L}anguage \textbf{A}lignment), a unified framework for efficient and semantically consistent multimodal remote sensing understanding. Specifically, we introduce a Dynamic Resolution Input Strategy (\textbf{DRIS}), which adaptively allocates computational resources and visual tokens according to scene complexity, enabling efficient visual perception while preserving fine-grained semantic details. Building upon these resolution-adaptive representations, we further develop a Multi-scale Vision--Language Alignment Mechanism (\textbf{MS-VLAM}), which decomposes cross-modal alignment into object-level, region-level, and global-level semantic spaces to establish hierarchical semantic correspondence across different semantic granularities. By jointly integrating adaptive visual representation learning and hierarchical semantic alignment, RAH-VLA effectively bridges the gap between visual representations and semantic understanding in remote sensing imagery, providing a unified framework for image captioning, visual grounding, and multimodal reasoning.

The main contributions of this work are summarized as follows:

\begin{enumerate}

\item We propose \textbf{RAH-VLA} (Resolution-Adaptive Hierarchical Vision--Language Alignment), a unified framework for multimodal remote sensing understanding that jointly models adaptive visual perception and hierarchical semantic alignment.

\item We develop a Dynamic Resolution Input Strategy (\textbf{DRIS}) that adaptively allocates visual computation according to scene complexity, reducing computational redundancy while preserving fine-grained visual details.

\item We introduce a Multi-scale Vision--Language Alignment Mechanism (\textbf{MS-VLAM}) to establish hierarchical semantic correspondence across object-level, region-level, and global-level representations.

\item Extensive experiments on multiple remote sensing benchmarks demonstrate the effectiveness of RAH-VLA in image captioning, visual grounding, and multimodal reasoning tasks, while additional analyses validate its interpretability and cross-modality generalization capability.

\end{enumerate}

\section{Related Work}

\subsection{Multimodal Semantic Understanding in Remote Sensing}

The rapid growth of heterogeneous Earth observation data has driven increasing interest in multimodal remote sensing (RS) semantic understanding, where complementary information from optical imagery, synthetic aperture radar (SAR), LiDAR, multispectral, and hyperspectral modalities is jointly exploited \cite{ref28,ref9}. Compared with single-modality analysis, multimodal frameworks integrate complementary spatial, spectral, geometric, and structural information across different sensing mechanisms, significantly improving robustness and scene understanding capability under complex observation conditions, including noise interference, cloud contamination, and heterogeneous terrain distributions \cite{ref24,ref13}. Consequently, multimodal semantic understanding has gradually evolved into a fundamental research direction for intelligent remote sensing analysis \cite{ref66}.

\textbf{Early shallow fusion stage.}
Early multimodal RS methods mainly relied on handcrafted features and shallow fusion strategies, where different modalities were combined at the pixel, feature, or decision levels \cite{ref29,ref30}. Traditional remote sensing analysis heavily depended on manually designed texture, shape, and spectral priors to characterize heterogeneous observations. Beyond handcrafted feature extraction, traditional hyperspectral image processing further explored spectral information exploitation through techniques such as image compression and spectral unmixing, aiming to improve data representation and reveal intrinsic spectral characteristics of hyperspectral imagery \cite{ref80,ref81}. Representative handcrafted descriptors include SIFT \cite{ref67}, HOG \cite{ref68}, LBP \cite{ref69}, and spectral indices \cite{ref70}. Although these methods improve multimodal information utilization to some extent, they implicitly assume that heterogeneous modalities can be directly projected into compatible feature spaces through shallow aggregation. However, substantial differences in sensing mechanisms, statistical distributions, and semantic characteristics among heterogeneous modalities often lead to severe modality gaps and inconsistent cross-modal semantics in remote sensing scenarios \cite{ref26}. Consequently, shallow fusion methods generally suffer from limited semantic interpretability, weak robustness to modality-specific noise, and poor generalization capability across complex remote sensing environments.

The limitations of shallow fusion methods in modeling structured multimodal interaction have driven the development of deep learning-based architectures for more effective contextual understanding.

\textbf{CNN-based multimodal interaction paradigms.}
In the early deep learning era, convolutional neural network(CNN)-based fusion networks became the dominant paradigm for multimodal RS analysis by replacing handcrafted features with automatically learned hierarchical features. Representative architectures typically adopted dual-stream or multi-branch CNN structures to extract modality-specific features, followed by multimodal interaction through feature concatenation, attention weighting, or multi-scale aggregation mechanisms \cite{ref72}. For example, Guo et al. proposed a CNN-based spatial feature fusion framework for hyperspectral image classification \cite{ref32}, while Feng et al. developed ICIF-Net to model intra-scale cross-modal interaction and inter-scale feature fusion for complex remote sensing scenes \cite{ref33}.

Compared with shallow fusion methods, CNN-based architectures significantly improve local contextual perception and multimodal interaction through hierarchical feature extraction. However, due to the localized nature of convolutional operations, these methods mainly capture short-range neighborhood dependencies and struggle to model long-range spatial interaction in large-scale remote sensing scenes \cite{ref73}. Moreover, semantic interaction is typically performed through implicit fusion modules without explicitly constraining structured cross-modal consistency across heterogeneous modalities.

The limitations of CNN-based methods in modeling global contextual interaction have further motivated the introduction of Transformer-based architectures.

\textbf{Transformer-based multimodal interaction paradigms.}
Transformer-based architectures introduce self-attention mechanisms to model long-range dependencies and global contextual interaction across heterogeneous modalities. Representative models such as ViT \cite{ref34} and Swin Transformer \cite{ref61} significantly improve global contextual modeling and multimodal interaction in remote sensing scenarios. Building upon these architectures, recent studies further extend Transformer-based frameworks toward multimodal remote sensing understanding through large-scale contextual modeling and cross-modal attention mechanisms, while hybrid CNN-Transformer architectures are also explored to jointly capture local spatial details and global semantic dependencies \cite{ref74}.

However, despite their stronger global modeling capability, most existing Transformer-based methods still mainly rely on implicit feature interaction and coarse feature-level fusion without explicitly modeling hierarchical semantic structures or structured cross-modal consistency under heterogeneous sensing conditions. Consequently, consistent semantic interaction across multiple spatial granularities remains insufficiently explored in complex remote sensing scenarios.

These limitations have recently driven the development of scalable multimodal foundation paradigms for more generalized remote sensing understanding.

\textbf{Recent multimodal foundation paradigms.}
Building upon Transformer architectures and large-scale self-supervised learning, recent research has gradually shifted toward scalable multimodal foundation paradigms for increasingly large-scale and heterogeneous remote sensing data \cite{ref24}. Multimodal self-supervised learning enables transferable semantic interaction across heterogeneous sensing modalities by leveraging large-scale unlabeled RS data \cite{ref28}. Representative studies further incorporate geospatial relations, contrastive learning, and multimodal semantic interaction into self-supervised remote sensing frameworks \cite{ref35,ref36}.

Recent RS foundation models further improve multimodal understanding capability through large-scale Earth observation pretraining, while semantic reasoning and semantic reconstruction mechanisms are increasingly introduced to strengthen multimodal interaction and contextual consistency \cite{ref37,ref39,ref40,ref78}.

Nevertheless, most existing methods still rely on implicit feature interaction and coarse feature-level fusion without explicitly modeling hierarchical semantic correspondence across heterogeneous modalities \cite{ref41}. Consequently, robust multimodal understanding across multiple spatial granularities remains insufficiently explored, particularly under extreme scale variation and spatially distributed semantics. These challenges have motivated the development of more adaptive and structured multimodal interaction paradigms for remote sensing understanding.

\subsection{Vision-Language Modeling for Remote Sensing}

Vision-language modeling has emerged as a dominant paradigm for unifying visual perception and natural language understanding through aligned visual--textual representations. In natural image domains, large-scale VLMs such as contrastive language-image pre-training(CLIP), foundational language and vision alignment(FLAVA), and CoCa demonstrate that pretraining on web-scale image--text data can produce highly transferable multimodal representations, enabling strong performance in image captioning, visual question answering, and cross-modal retrieval \cite{ref17,ref42,ref43}. Such success largely benefits from the intrinsic characteristics of natural images, where semantic concepts are typically object-centric, spatially compact, and strongly correlated with textual descriptions.

\textbf{General VLM paradigm and downstream extension.}
Driven by the increasing demand for intelligent remote sensing interpretation, recent studies have gradually extended general vision--language paradigms toward remote sensing (RS) scenarios \cite{ref5}. Unlike conventional visual recognition tasks that mainly focus on category-level recognition, RS-oriented VLMs require joint understanding of complex spatial layouts, structural relationships, and contextual scene information for downstream multimodal reasoning tasks \cite{ref11}. Consequently, RS VLMs have been increasingly applied to image captioning, visual grounding, and multimodal reasoning tasks \cite{ref44}.

Recent RS-oriented VLM frameworks further enhance multimodal interaction by incorporating geospatial grounding, large-scale domain-adaptive pretraining, and remote sensing structural priors into vision--language architectures. Representative examples such as GeoChat and remote sensing GPT(RSGPT) demonstrate the feasibility of adapting large-scale vision--language models to complex remote sensing scenarios through grounded semantic interaction and domain-specific multimodal pretraining \cite{ref18,ref45,ref79}. The rapid emergence of these RS-oriented VLM frameworks has also driven increasing efforts toward systematic reviews of datasets, architectures, and multimodal reasoning paradigms for remote sensing vision--language modeling \cite{ref20,ref24}.

\textbf{Paradigm mismatch between conventional VLMs and remote sensing imagery.}
Despite their promising capability, conventional VLMs are primarily developed under natural-image assumptions, where visual semantics are usually concentrated on a few salient foreground objects with relatively compact spatial distributions. However, such assumptions are fundamentally violated in remote sensing scenarios due to the intrinsic properties of RS imagery.

Unlike natural images, RS scenes exhibit extreme scale variation, where large homogeneous land-cover regions coexist with small but visually critical objects, making uniform feature extraction pipelines ineffective for simultaneously preserving global contextual information and fine-grained local details \cite{ref25}. Moreover, semantic information in RS imagery is inherently spatially distributed, where meaningful scene understanding often depends on complex interactions among multiple regions rather than isolated object-centric semantics \cite{ref48}. In addition, the correspondence between visual patterns and textual descriptions is frequently weak and implicitly correlated, further increasing the difficulty of precise visual--textual interaction and semantic correspondence modeling \cite{ref49}.

Consequently, directly transferring conventional VLM paradigms to remote sensing scenarios often leads to insufficient visual--textual correspondence and unstable multimodal interaction under complex spatial distributions. These limitations reveal a fundamental mismatch between resolution-dependent visual encoding and hierarchically distributed remote sensing semantics.

\textbf{Limitations of existing RS VLM methods.}
To alleviate the above challenges, existing RS VLM approaches mainly focus on enhancing visual representation capability and improving multimodal interaction mechanisms \cite{ref24}. Nevertheless, most existing methods still rely on fixed-resolution visual encoding pipelines and single-level global interaction strategies, which are inherently incompatible with the hierarchical semantic structures and multi-scale spatial characteristics of RS scenes \cite{ref50}. Fixed-resolution visual encoding often introduces a trade-off between computational efficiency and fine-grained detail preservation, while global-only interaction mechanisms fail to explicitly model semantic correspondence across different spatial granularities \cite{ref27}.

Moreover, semantic interaction in most existing RS VLMs is predominantly performed at a single semantic granularity, making it difficult to establish consistent correspondence across object-level, regional-level, and scene-level semantics. Consequently, current RS VLMs struggle to simultaneously preserve local fine-grained details and maintain globally coherent contextual understanding, resulting in suboptimal multimodal interaction performance under complex remote sensing scenarios.

These observations suggest that effective RS vision--language modeling requires moving beyond fixed-resolution encoding and coarse global interaction toward more adaptive and structured multimodal semantic modeling paradigms, which remain insufficiently explored in existing research. These limitations motivate the development of RAH-VLA, which provides a unified framework for jointly modeling resolution-adaptive visual perception and hierarchical vision--language semantic alignment in multimodal remote sensing understanding.

\begin{figure*}[t]
    \centering
    \includegraphics[width=0.9\textwidth]{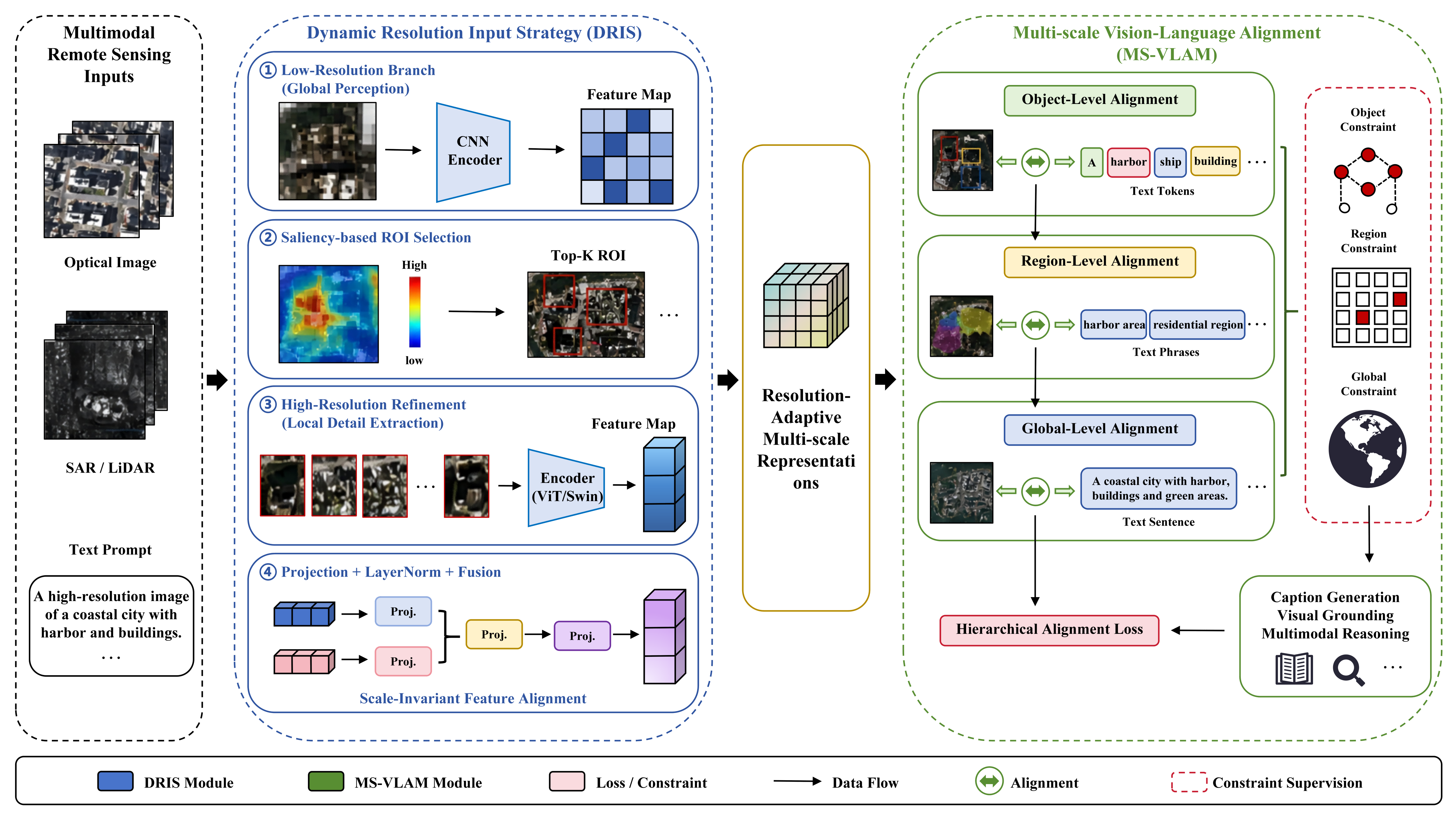} 
    \caption{
    \textbf{Overview of the proposed RAH-VLA framework for multimodal remote sensing interpretation.}
    DRIS combines low-resolution global perception and high-resolution local refinement for adaptive resource allocation. MS-VLAM performs hierarchical cross-modal alignment at object, region and global levels. The framework is jointly optimized by hierarchical alignment loss.
}
    \label{fig:model_framework1}
\end{figure*}

\section{Method}

\subsection{Overview of the Proposed Framework}

This study proposes RAH-VLA, a resolution-adaptive hierarchical vision--language alignment framework for multimodal remote sensing understanding, as illustrated in Fig.~\ref{fig:model_framework1}. The framework is designed to address two key challenges in RS vision--language modeling: scale inconsistency caused by fixed-resolution visual processing, and semantic mismatch caused by single-level vision--language alignment. For ease of reference, the major mathematical symbols and variables used throughout this section are summarized in Table~\ref{tab:notation}. 

\begin{table}[t]
\centering
\caption{Summary of Major Mathematical Notations}
\label{tab:notation}
\renewcommand{\arraystretch}{1.1}
\setlength{\tabcolsep}{3pt}
\begin{tabular}{lll}
\toprule
\textbf{Symbol} & \textbf{Description} & \textbf{Eq.} \\
\midrule
$R(x,y)$ & Resolution assignment function & (1) \\
$R_{\rm high},R_{\rm low}$ & High-/low-resolution modes & (1) \\
$s(x,y)$ & Regional saliency score & (1) \\
$\tau$ & Adaptive threshold value & (1)--(2) \\
$\lambda$ & Threshold sensitivity coefficient & (2) \\
$k$ & Number of selected ROIs & (6) \\
$\Omega$ & Selected region of interest & (3)--(5) \\
$F_{\rm coarse}$ & Coarse-resolution feature map & (3)--(5) \\
$F_{\rm fine}$ & Fine-resolution feature map & (3)--(5) \\
$F_{\rm fused}$ & Fused multi-resolution features & (3)--(5) \\
$P_{\rm ROI}$ & Selected ROI set & (6) \\
$V$ & Visual feature representation & (7), (17) \\
$g$ & Global visual representation & (17)--(18) \\
$t_{\rm CLS}$ & Global textual representation & (18) \\
$L_{\rm obj}$ & Object-level alignment loss & (10) \\
$L_{\rm reg}$ & Region-level alignment loss & (16) \\
$L_{\rm glob}$ & Global-level alignment loss & (18) \\
$\alpha,\beta,\gamma$ & Alignment loss weights & (19) \\
$L_{\rm align}$ & Hierarchical alignment loss & (19) \\
$L_{\rm caption}$ & Caption generation loss & (20) \\
$\delta$ & Loss balancing coefficient & (21) \\
$L$ & Overall training objective & (21) \\
\bottomrule
\end{tabular}
\end{table}

As shown in the figure, the framework consists of two closely coupled core mechanisms, which implement the four basic components of a typical vision--language model: the \textbf{Dynamic Resolution Input Strategy (DRIS)} serves as the visual encoder, and the \textbf{Multi-scale Vision--Language Alignment Mechanism (MS-VLAM)} integrates the connector, text encoder, and alignment functions.

Given an RS image and its corresponding textual prompt, the DRIS visual encoder first processes the image in a coarse-to-fine manner: it performs low-resolution global perception to estimate regional saliency, selects semantically important regions for high-resolution refinement, and then generates resolution-adaptive visual representations through cross-resolution feature fusion. Meanwhile, the text prompt is embedded into semantic vectors by the text encoder. The connector module (implemented within MS-VLAM) maps visual features into a language-compatible space, enabling the framework to perform hierarchical cross-modal alignment at object-level, region-level, and global-level granularities. Finally, the aligned vision--language representations are fed into the large language model (LLM) to support downstream multimodal understanding tasks.

\subsection{Relationship between DRIS and MS-VLAM}

\begin{table*}
\centering
\caption{Comparison of representative vision--language frameworks in terms of adaptive visual perception, semantic alignment, and optimization strategy.}
\label{tab:comparison}
\renewcommand{\arraystretch}{1.15}
\setlength{\tabcolsep}{4pt}
\footnotesize
\begin{tabular}{lccccc}
\toprule
\textbf{Design Aspect} &
\textbf{DynamicViT} &
\textbf{Token} &
\textbf{GeoChat} &
\textbf{RSGPT} &
\textbf{RAH-VLA} \\
&
&
\textbf{Merging}
&
&
&
\\
\midrule
Adaptive Resolution        & $\times$ & $\times$ & $\times$ & $\times$ & $\checkmark$ \\
Token Selection            & $\checkmark$ & $\checkmark$ & $\times$ & $\times$ & $\checkmark$ \\
Object-level Alignment     & $\times$ & $\times$ & $\times$ & $\times$ & $\checkmark$ \\
Region-level Alignment     & $\times$ & $\times$ & $\times$ & $\times$ & $\checkmark$ \\
Global-level Alignment           & $\times$ & $\times$ & $\checkmark$ & $\checkmark$ & $\checkmark$ \\
Joint Optimization         & $\times$ & $\times$ & $\times$ & $\times$ & $\checkmark$ \\
\bottomrule
\end{tabular}
\end{table*}

Unlike conventional vision--language pipelines, where adaptive visual processing and cross-modal alignment are optimized independently, the proposed RAH-VLA establishes a unified framework that tightly couples the Dynamic Resolution Input Strategy (DRIS) and the Multi-scale Vision--Language Alignment Mechanism (MS-VLAM). The motivation behind this design is that remote sensing images simultaneously exhibit large-scale scene diversity and fine-grained object variations, making efficient visual perception and hierarchical semantic alignment inherently interdependent rather than isolated optimization objectives.

To better illustrate this design philosophy, Table~\ref{tab:comparison} compares representative vision--language frameworks with the proposed RAH-VLA. Existing adaptive visual processing approaches, such as DynamicViT and Token Merging, primarily improve computational efficiency through token selection or adaptive computation while lacking explicit cross-modal semantic modeling. In contrast, representative remote sensing vision--language models, including GeoChat and RSGPT, generally rely on fixed-resolution visual representations and perform semantic alignment mainly at the global level. Consequently, adaptive visual perception and hierarchical vision--language alignment are usually designed and optimized as separate components.

The proposed framework bridges this gap by enabling mutual interaction between DRIS and MS-VLAM. Specifically, DRIS is not merely introduced to reduce computational redundancy. By allocating high-resolution computation to semantically informative regions while preserving low-resolution global context, DRIS generates resolution-adaptive multi-scale visual representations with enhanced object boundaries, richer regional structures, and more coherent scene-level semantics. These structured representations provide informative visual inputs for subsequent hierarchical semantic alignment.

Building upon these adaptive visual representations, MS-VLAM establishes semantic correspondence at the object-, region-, and global-levels through hierarchical cross-modal supervision. Rather than treating visual encoding and semantic alignment as two independent stages, the hierarchical alignment objective continuously refines the visual representations produced by DRIS during training. Consequently, DRIS improves the quality of visual features for semantic alignment, whereas MS-VLAM further enhances their semantic consistency through multi-level supervision.

The two modules are jointly optimized in an end-to-end manner, allowing adaptive visual perception and hierarchical semantic alignment to mutually reinforce each other throughout the learning process. This unified optimization strategy enables the proposed framework to simultaneously improve computational efficiency, preserve fine-grained visual details, and achieve more consistent cross-modal semantic understanding for multimodal remote sensing tasks.

\subsection{Dynamic Resolution Input Strategy (DRIS)}

\begin{figure*}[htbp]
    \centering
    \includegraphics[width=0.9\textwidth]{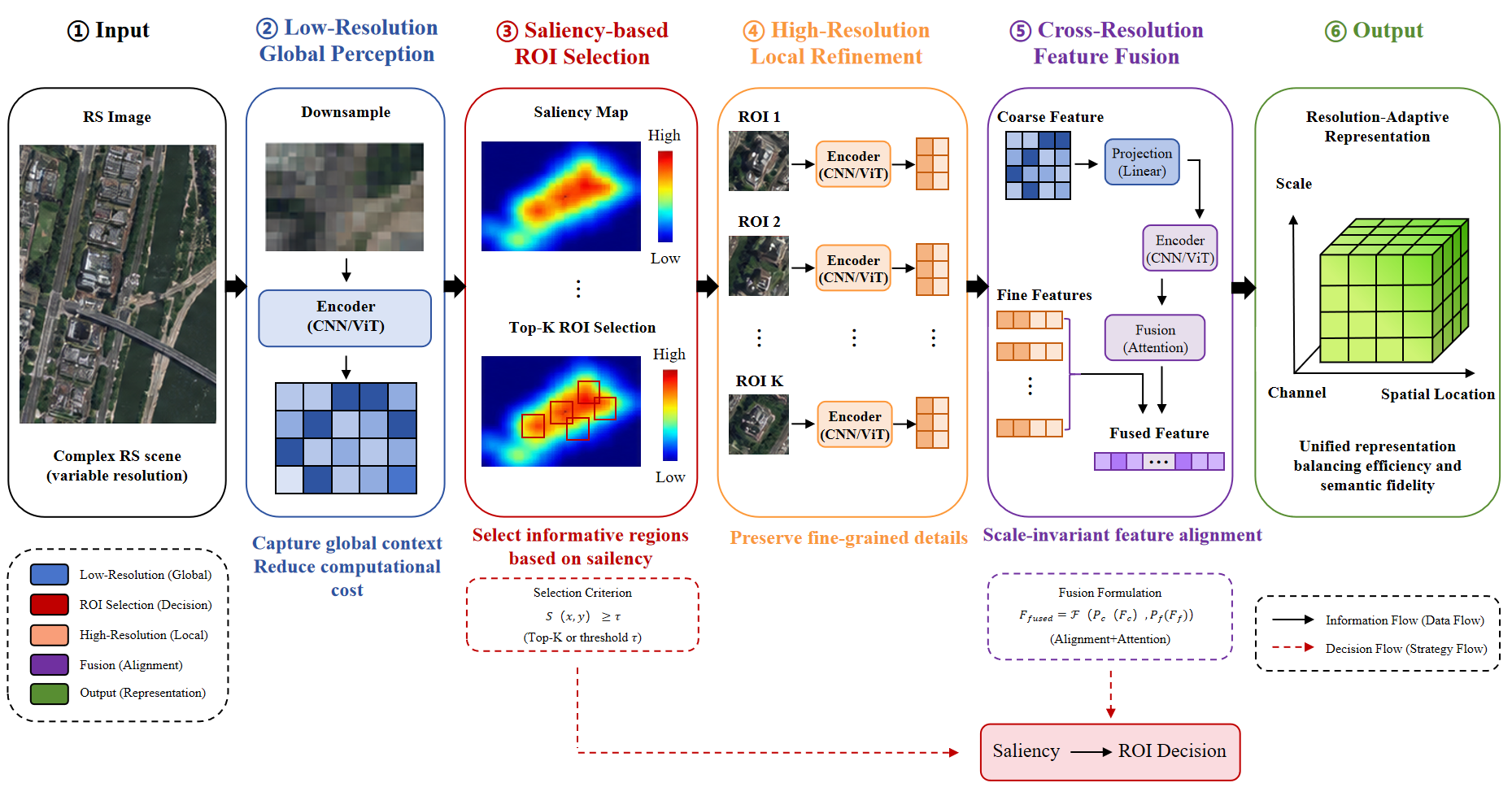}
    \caption{
      \textbf{Illustration of the proposed Dynamic Resolution Input Strategy (DRIS).}
      The DRIS adopts a coarse-to-fine, resolution-adaptive paradigm. It first captures global context via low-resolution perception and generates saliency maps for ROI selection. Then, selected regions are refined at high resolution to preserve fine details. Cross-resolution feature fusion finally produces resolution-adaptive representations balancing semantic fidelity and computational efficiency.
}
    \label{fig:model_framework2}
\end{figure*}

Remote sensing images usually cover large spatial areas and contain complex land-cover structures. Directly using fixed high-resolution inputs leads to excessive GPU memory consumption and computational cost, whereas fixed low-resolution inputs may lose small objects, boundary information, and texture details. Consequently, existing fixed-resolution processing strategies often suffer from an inherent trade-off between computational efficiency and fine-grained semantic perception.

To address this limitation, we propose a Dynamic Resolution Input Strategy (DRIS), which dynamically allocates computational resources according to scene complexity and semantic importance. As illustrated in Fig.~\ref{fig:model_framework2}, DRIS adopts a coarse-to-fine resolution allocation paradigm. Specifically, a low-resolution branch is first employed to obtain global scene understanding and estimate regional saliency. Subsequently, semantically important regions are selected for high-resolution refinement, and cross-resolution feature fusion is performed to generate resolution-adaptive visual representations.

The dynamic resolution allocation function is defined as:
\begin{equation}
R(x,y)=
\begin{cases}
R_{\text{high}}, & \text{if } s(x,y) \ge \tau, \\
R_{\text{low}}, & \text{otherwise},
\end{cases}
\tag{1}
\end{equation}
where $R(x,y)$ denotes the resolution mode assigned to spatial position $(x,y)$; $R_{\text{high}}$ and $R_{\text{low}}$ denote the high-resolution and low-resolution processing modes, respectively; $s(x,y)$ denotes the regional significance score estimated from the low-resolution global analysis; and $\tau$ is the threshold used to determine whether a region should be processed at high resolution. When $s(x,y)\ge\tau$, the corresponding region is regarded as semantically salient or structurally complex and is processed using $R_{\text{high}}$; otherwise, it is processed using $R_{\text{low}}$.

To make the resolution allocation adaptive to different RS scenes, the threshold $\tau$ is further defined as:
\begin{equation}
\tau = \mu_s + \lambda \cdot \sigma_s,
\tag{2}
\end{equation}
where $\mu_s$ and $\sigma_s$ denote the mean and standard deviation of the saliency scores in the coarse-resolution attention map, respectively, and $\lambda$ is a sensitivity coefficient controlling the strictness of high-resolution region selection. A larger $\lambda$ selects fewer but more salient regions, while a smaller $\lambda$ allows more regions to enter the high-resolution branch. In all experiments, $\lambda$ is set to 1.0 unless otherwise stated. This adaptive threshold design enables DRIS to dynamically adjust resolution allocation according to scene complexity.

In practice, the adaptive threshold first identifies candidate salient regions according to scene-dependent saliency statistics. To maintain a fixed computational budget across different scenes, a Top-$k$ selection strategy is subsequently applied to the candidate set, and the $k$ highest-scoring regions are selected for high-resolution refinement. Therefore, the adaptive threshold is responsible for candidate region generation, whereas the Top-$k$ operation determines the final ROIs used for local detail extraction. In all experiments, $k$ is fixed to 5 to ensure consistent computational complexity, stable training behavior, and fair comparison across different evaluation settings.

The mechanism first performs low-resolution processing on a large-format RS image to obtain global scene understanding, coarse localization, and regional saliency estimation. If the downsampling factor is $n$, the computational complexity of low-resolution global processing is approximately $\frac{1}{n^2}$ of full-resolution processing. Subsequently, high-resolution image blocks are extracted only from selected regions of interest (ROIs), enabling fine-grained analysis of small targets, boundary structures, and complex textures.

To integrate features from different resolutions, a multi-resolution feature fusion strategy is adopted. Let $F_{\text{coarse}}$ denote the low-resolution feature map extracted from the global view, and let $F_{\text{fine}}$ denote the high-resolution localized feature map extracted from selected ROIs. The initial fusion process can be expressed as:
\begin{equation}
F_{\text{fused}} =
\text{Upsample}\left(F_{\text{coarse}}\right)
+
\text{Crop}\left(F_{\text{fine}}, \Omega\right),
\tag{3}
\end{equation}
where $F_{\text{fused}}$ denotes the fused multi-resolution feature map; $\Omega$ denotes the spatial range of the selected ROI; $\text{Upsample}(\cdot)$ denotes the operation that resizes the coarse feature map to match the target spatial scale; and $\text{Crop}(\cdot,\Omega)$ denotes the operation that extracts the high-resolution localized feature region corresponding to $\Omega$. This operation aims to combine global contextual information from the low-resolution branch with fine-grained structural details from the high-resolution branch.

However, $F_{\text{coarse}}$ and $F_{\text{fine}}$ originate from different resolutions and receptive fields, and therefore may exhibit different statistical distributions. Directly adding them may introduce feature inconsistency across scales. To address this issue, both feature maps are first projected into a shared latent space and normalized before fusion:
\begin{equation}
\tilde{F}_{\text{coarse}}
=
\text{LN}\left(W_c F_{\text{coarse}}\right),
\quad
\tilde{F}_{\text{fine}}
=
\text{LN}\left(W_f F_{\text{fine}}\right),
\tag{4}
\end{equation}
where $\tilde{F}_{\text{coarse}}$ and $\tilde{F}_{\text{fine}}$ denote the normalized coarse and fine feature maps after projection; $W_c$ and $W_f$ are learnable projection matrices for coarse and fine features, respectively; and $\text{LN}(\cdot)$ denotes Layer Normalization. The final aligned feature fusion is then formulated as:
\begin{equation}
F_{\text{fused}} =
\text{Upsample}\left(\tilde{F}_{\text{coarse}}\right)
+
\text{Crop}\left(\tilde{F}_{\text{fine}}, \Omega\right).
\tag{5}
\end{equation}
Before fusion, the projected coarse- and fine-resolution features are resized to the same spatial and channel dimensions, ensuring compatibility for cross-resolution aggregation. This design can be interpreted as a projection-based cross-resolution feature alignment process. By mapping coarse and fine features into a shared latent space before fusion, the model preserves global semantic context from low-resolution inputs while incorporating fine-grained details from high-resolution regions, thereby alleviating semantic discontinuity caused by direct cross-scale aggregation.

Furthermore, the projection layers implicitly learn a scale-invariant embedding space, in which features from different resolutions are aligned according to semantic similarity rather than spatial scale. This enables the fused representation to maintain semantic consistency while integrating multi-scale information, thus mitigating potential discontinuities caused by heterogeneous receptive fields.

For ROI selection, an attention heatmap is generated in the low-resolution branch:
\begin{equation}
P_{\text{ROI}} =
\text{Topk}\left(
\text{GaussianBlur}\left(A_{\text{coarse}}\right), k
\right),
\tag{6}
\end{equation}
where $P_{\text{ROI}}$ denotes the selected ROI set; $A_{\text{coarse}} \in \mathbb{R}^{\frac{H}{n}\times\frac{W}{n}}$ denotes the coarse attention heatmap generated from the low-resolution representation; $H$ and $W$ denote the height and width of the original image; $n$ denotes the downsampling factor; $\text{GaussianBlur}(\cdot)$ denotes the smoothing operation used to suppress noisy responses and preserve stable salient regions; and $\text{Topk}(\cdot,k)$ selects the top-$k$ regions with the highest response values as candidate ROIs. The Gaussian smoothing parameter $\sigma$ controls the trade-off between noise suppression and boundary preservation.

Through this design, DRIS reduces redundant computation and improves inference efficiency while preserving fine-grained visual details. More importantly, the resolution-adaptive features generated by DRIS provide structured visual inputs for the subsequent MS-VLAM alignment mechanism. Therefore, DRIS is not merely an efficiency-oriented preprocessing strategy, but also a representation mechanism that supports hierarchical vision--language alignment. More importantly, the enhanced object-aware and region-aware representations generated by DRIS directly provide reliable multi-scale visual cues for the subsequent MS-VLAM, facilitating more accurate hierarchical semantic alignment across different spatial granularities.

\subsection{Multi-scale Vision--Language Alignment Mechanism (MS-VLAM)}

\begin{figure*}[htbp]
    \centering
    \includegraphics[width=0.9\textwidth]{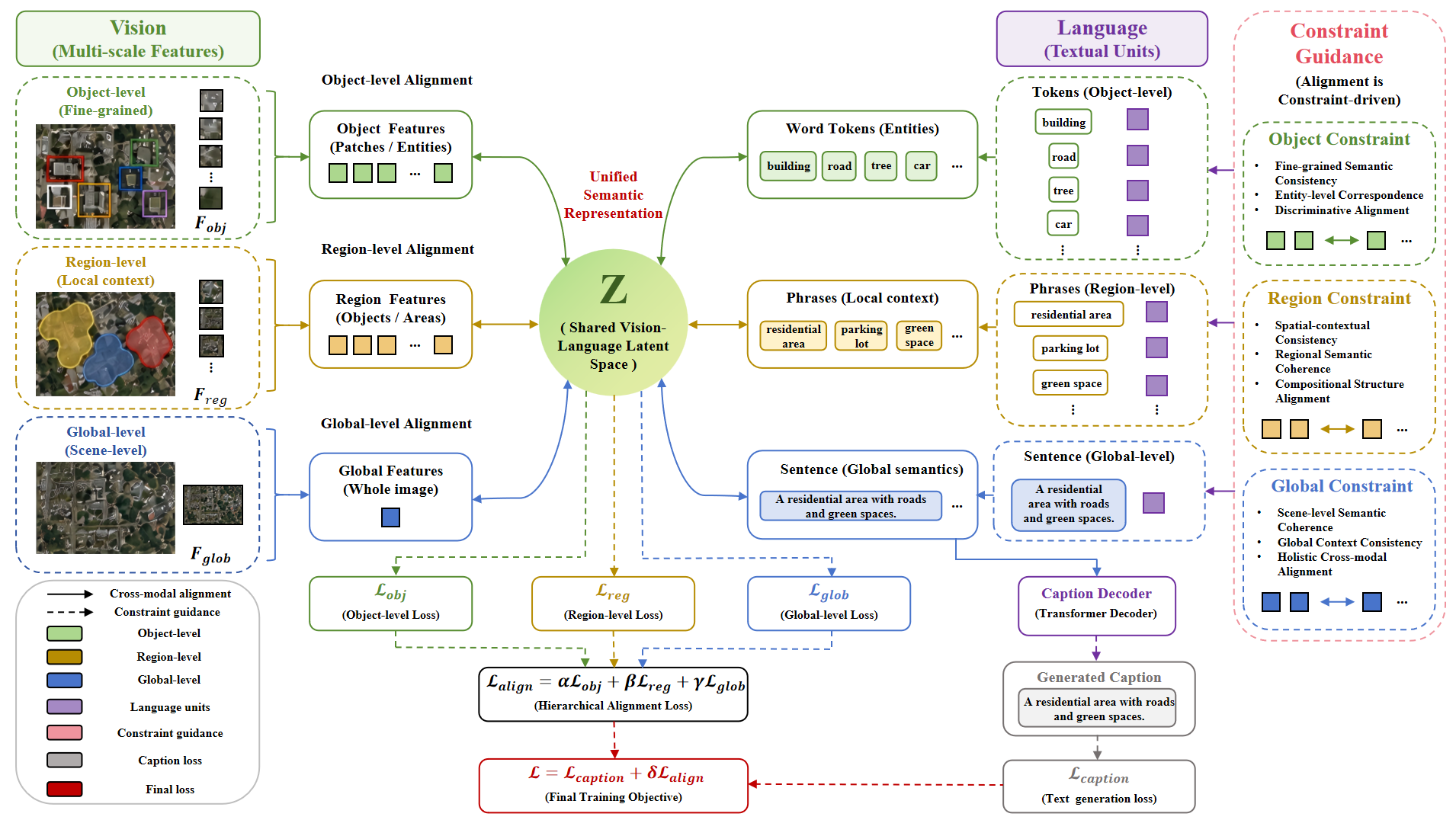}
    \caption{
      \textbf{Overview of the proposed Multi-scale Vision--Language Alignment Mechanism (MS-VLAM).}
      MS-VLAM performs hierarchical cross-modal alignment between multi-scale visual representations and multi-granularity textual semantics. Object-, region-, and global-level representations are aligned within a shared vision--language latent space under hierarchical semantic constraints. The resulting alignment objective jointly optimizes semantic consistency and caption generation.
    }
    \label{fig:model_framework3}
\end{figure*}

Building upon the resolution-adaptive visual representations generated by DRIS, MS-VLAM further organizes these adaptive features into object-level, region-level, and global-level semantic representations. Since the informative regions have already been selectively enhanced by DRIS, hierarchical alignment can focus on semantically important structures rather than redundant background information, resulting in more effective cross-modal correspondence. Existing vision--language models mainly rely on global or single-scale alignment strategies, which often match an entire image with a sentence-level representation. Such a design is insufficient for remote sensing imagery, where semantic information is inherently distributed across multiple spatial granularities, ranging from individual objects and local regions to holistic scene semantics. Consequently, semantic mismatch may occur when object-level details, regional context, and global scene understanding need to be jointly modeled.

To address this limitation, we propose a Multi-scale Vision--Language Alignment Mechanism (MS-VLAM), which explicitly performs hierarchical cross-modal alignment across multiple semantic levels. As illustrated in Fig.~\ref{fig:model_framework3}, MS-VLAM decomposes vision--language interaction into object-level, region-level, and global-level semantic spaces, enabling structured semantic correspondence between resolution-adaptive visual representations and textual semantics at different granularities.

The object-level alignment focuses on fine-grained entities such as buildings, aircraft, vehicles, ships, and roads. The region-level alignment models local spatial contexts and interactions among objects, such as roads adjacent to water bodies or buildings surrounded by vegetation. The global-level alignment captures holistic scene semantics, such as airports, ports, residential areas, or agricultural regions. By jointly optimizing these three levels, MS-VLAM improves fine-grained semantic correspondence while preserving global contextual coherence.

Concretely, the framework first extracts salient object representations from region proposals generated by the visual encoder and RoI pooling, then aligns them with entity-level textual features. Meanwhile, semantic masks or segmented regions are used to generate region-level visual features, which are matched with phrase-level textual embeddings. Finally, global visual representations are derived using Spatial Pyramid Pooling (SPP) and aligned with sentence-level textual representations. The three alignment losses are then combined to support robust caption generation, visual grounding, and multimodal reasoning.

\subsubsection{Object-level Alignment Loss}

At the object level, visual features are extracted from the $p^{\text{th}}$ candidate object region identified from the visual feature map through region proposal generation and RoI pooling. The visual feature vector of the $p^{\text{th}}$ object is defined as:
\begin{equation}
\mathbf{v}_{\text{obj}}^{(p)}
=
f_{\text{v}}\left(\text{RoI}(V,\mathbf{B}_p)\right),
\tag{7}
\end{equation}
where $V$ denotes the overall visual feature map of the input image; $\mathbf{B}_p$ denotes the bounding box of the $p^{\text{th}}$ candidate object; $\text{RoI}(\cdot)$ denotes RoI pooling or RoIAlign, which extracts local visual features from $V$ according to $\mathbf{B}_p$; and $f_{\text{v}}(\cdot)$ denotes a visual projection function, such as a fully connected layer or linear transformation, that maps RoI features into the shared alignment space.

The corresponding text entity feature is defined as:
\begin{equation}
\mathbf{o}_{\text{obj}}^{(p)}
=
f_{\text{t}}(\mathbf{e}_p),
\tag{8}
\end{equation}
where $\mathbf{e}_p$ denotes the textual representation of the corresponding entity, such as an object category label or descriptive phrase, and $f_{\text{t}}(\cdot)$ denotes the text projection function that maps entity-level text features into the same alignment space as visual object features.

The semantic similarity between visual and textual features is measured using cosine similarity:
\begin{equation}
\cos(\mathbf{u},\mathbf{v})
=
\frac{\mathbf{u}^{\top}\mathbf{v}}
{\|\mathbf{u}\|\|\mathbf{v}\|},
\tag{9}
\end{equation}
where $\mathbf{u}^{\top}\mathbf{v}$ denotes the inner product between vectors $\mathbf{u}$ and $\mathbf{v}$, and $\|\mathbf{u}\|$ and $\|\mathbf{v}\|$ denote their Euclidean norms.

Considering that object localization quality affects alignment reliability, an IoU-based dynamic weighting strategy is introduced. The object-level alignment loss is formulated as:
\begin{equation}
\mathcal{L}_{\text{obj}}
=
1-\frac{1}{P}
\sum_{p=1}^{P}
w_p \cdot
\cos\left(
\mathbf{v}_{\text{obj}}^{(p)},
\mathbf{o}_{\text{obj}}^{(p)}
\right),
\tag{10}
\end{equation}
where $P$ denotes the number of object instances; $\mathbf{v}_{\text{obj}}^{(p)}$ and $\mathbf{o}_{\text{obj}}^{(p)}$ denote the visual and textual features of the $p^{\text{th}}$ object, respectively; and $w_p$ denotes the localization-aware weight assigned to this object.

The weight $w_p$ is computed as:
\begin{equation}
w_p =
\frac{
\text{IoU}\left(\hat{\mathbf{B}}_p,\mathbf{B}_p\right)
}{
\sum_{q=1}^{P}
\text{IoU}\left(\hat{\mathbf{B}}_q,\mathbf{B}_q\right)
},
\tag{11}
\end{equation}
where $\hat{\mathbf{B}}_p$ denotes the predicted bounding box of the $p^{\text{th}}$ object, $\mathbf{B}_p$ denotes the corresponding reference bounding box, and $\text{IoU}(\cdot)$ denotes the intersection-over-union metric. This weighting strategy strengthens the contribution of accurately localized objects and improves the stability of fine-grained semantic alignment.

\subsubsection{Region-level Alignment Loss}

At the region level, semantic masks or segmented regions generated by Segment Anything Model(SAM) or comparable segmentation models are used as candidate regions $\{R_k\}_{k=1}^{K}$, where $K$ denotes the number of regions. Each region undergoes masked pooling or masked RoI operation to obtain the region-level visual feature $\mathbf{v}_k$. On the textual side, a phrase extractor generates phrase candidates $\{t_j\}_{j=1}^{M}$, where $M$ denotes the number of textual phrases. Each phrase is mapped into a phrase-level embedding:
\begin{equation}
\mathbf{p}_j =
\text{PhraseEmbed}(t_j),
\tag{12}
\end{equation}
where $\mathbf{p}_j$ denotes the embedding of the $j^{\text{th}}$ phrase, and $\text{PhraseEmbed}(\cdot)$ denotes the phrase embedding function.

Since the correspondence between regions and phrases is often one-to-many or many-to-one, direct one-to-one hard matching may not be robust. Therefore, we combine hard region--phrase matching and contrastive soft matching. The hard matching loss is defined as:
\begin{equation}
\mathcal{L}_{\text{reg}}^{\text{hard}}
=
1-\frac{1}{K}
\sum_{k=1}^{K}
\cos\left(
\mathbf{v}_k,
\mathbf{p}_{\pi(k)}
\right),
\tag{13}
\end{equation}
where $\pi(k)$ denotes the phrase index assigned to the $k^{\text{th}}$ visual region.

For contrastive region--phrase alignment, the similarity score between the $k^{\text{th}}$ region and the $j^{\text{th}}$ phrase is computed as:
\begin{equation}
s_{kj}
=
\frac{\mathbf{v}_k^{\top}\mathbf{p}_j}{\tau_c},
\tag{14}
\end{equation}
where $s_{kj}$ denotes the similarity score, and $\tau_c>0$ denotes the temperature parameter for contrastive learning. Here, $\tau_c$ is used to distinguish it from the DRIS threshold $\tau$ in Eq.~(1).

The InfoNCE-based region-level loss is defined as:
\begin{equation}
\mathcal{L}_{\text{reg}}^{\text{NCE}}
=
-\frac{1}{K}
\sum_{k=1}^{K}
\log
\frac{
\exp\left(s_{k,y(k)}\right)
}{
\sum_{j=1}^{M}\exp\left(s_{kj}\right)
},
\tag{15}
\end{equation}
where $y(k)$ denotes the positive phrase index corresponding to the $k^{\text{th}}$ region.

The final region-level alignment loss is:
\begin{equation}
\mathcal{L}_{\text{reg}}
=
\mu \mathcal{L}_{\text{reg}}^{\text{hard}}
+
(1-\mu)\mathcal{L}_{\text{reg}}^{\text{NCE}},
\quad
\mu\in[0,1],
\tag{16}
\end{equation}
where $\mu$ controls the balance between hard matching and contrastive soft matching. This formulation enables the model to capture region--phrase relationships and multi-object interactions more robustly.

\subsubsection{Global-level Alignment Loss}

At the global level, the overall scene representation is obtained by applying Spatial Pyramid Pooling (SPP) to the visual feature map $V$:
\begin{equation}
\mathbf{g} =
\text{SPP}(V),
\tag{17}
\end{equation}
where $\mathbf{g}$ denotes the global visual vector, and $\text{SPP}(\cdot)$ denotes the spatial pyramid pooling operation. In practice, SPP aggregates features using multiple pooling scales, such as $1\times1$, $2\times2$, and $4\times4$, to capture scene-level context at different spatial granularities.

The corresponding textual representation is the sentence-level vector $t_{[\text{CLS}]}$, which is obtained from the classification(CLS) token or an equivalent global textual embedding. The global-level alignment loss is defined as:
\begin{equation}
\mathcal{L}_{\text{glob}}
=
1-
\cos\left(
\mathbf{g},
t_{[\text{CLS}]}
\right),
\tag{18}
\end{equation}
where $\mathcal{L}_{\text{glob}}$ measures the semantic discrepancy between the global visual context and sentence-level textual semantics. Layer normalization and L2 normalization are applied before computing similarity to improve numerical stability.

The overall hierarchical alignment objective is formulated as:
\begin{equation}
\mathcal{L}_{\text{align}}
=
\alpha \mathcal{L}_{\text{obj}}
+
\beta \mathcal{L}_{\text{reg}}
+
\gamma \mathcal{L}_{\text{glob}},
\tag{19}
\end{equation}
where $\alpha$, $\beta$, and $\gamma$ are non-negative weights controlling the contributions of object-level, region-level, and global-level alignment losses, respectively. In all experiments, $\alpha$, $\beta$, and $\gamma$ are treated as learnable parameters and optimized jointly with the model. To ensure stable optimization and balanced contributions from different semantic levels, the weights are normalized through a softmax operation, satisfying
\[
\alpha+\beta+\gamma=1.
\]
This design improves training stability while enabling adaptive weighting across object-level, region-level, and global-level alignment objectives.

\subsection{Caption Generation Loss and Optimization}

To jointly optimize semantic alignment and language generation, the hierarchical alignment loss is combined with the caption generation loss. Given a ground-truth caption sequence $(x_1,x_2,\ldots,x_T)$, the caption generation loss is defined as:
\begin{equation}
\mathcal{L}_{\text{caption}}
=
-\sum_{t=1}^{T}
\log
P_{\text{L}}
\left(
x_t \mid V, x_{<t}
\right),
\tag{20}
\end{equation}
where $T$ denotes the length of the caption sequence; $x_t$ denotes the $t^{\text{th}}$ token; $x_{<t}$ denotes all previously generated tokens before time step $t$; $V$ denotes the visual representation used by the decoder; and $P_{\text{L}}(\cdot)$ denotes the probability predicted by the language model.

The final training objective is:
\begin{equation}
\mathcal{L}
=
\mathcal{L}_{\text{caption}}
+
\delta\mathcal{L}_{\text{align}},
\tag{21}
\end{equation}
where $\delta>0$ is a balancing coefficient that controls the contribution of hierarchical alignment to the overall optimization objective.

This joint objective enables the model to jointly optimize language generation and multi-level vision--language alignment, leading to semantically consistent multimodal remote sensing understanding. As a result, the learned representation is both resolution-adaptive and semantically consistent across object, region, and global levels.

\section{Dataset Description}

Remote sensing vision--language benchmarks provide the basis for training and evaluating multimodal remote sensing understanding models. In this work, the proposed framework is trained on the large-scale multi-task benchmark \textbf{RS-GPT4V}\cite{ref45}, which provides instruction-following supervision across diverse remote sensing vision--language tasks.

\textbf{RS-GPT4V} is a large-scale instruction-based remote sensing vision--language benchmark containing 91,937 training images with 991,206 instruction--answer pairs and 15,999 test images with 258,419 instruction--answer pairs. The benchmark supports a variety of remote sensing understanding tasks, including image captioning, visual question answering, detailed scene description, local-region description, and multi-turn conversation.

Compared with traditional single-task datasets, RS-GPT4V provides:

\begin{itemize}
\item \textbf{Large-scale data distribution}, which supports robust learning of resolution-adaptive visual perception in DRIS;
\item \textbf{Multi-level semantic supervision}, which aligns well with the hierarchical design of MS-VLAM;
\item \textbf{Multi-task learning capability}, which improves multimodal understanding and generalization performance.
\end{itemize}

To ensure fair and direct comparison with existing methods, image captioning performance is evaluated on the official test splits of four widely used benchmark datasets, including \textbf{remote sensing image captioning dataset(RSICD)}, \textbf{Northwestern Polytechnical University Captions(NWPU-Captions)}, \textbf{UC Merced-Captions(UCM-Captions)}, and \textbf{Sydney-Captions}.

For visual grounding evaluation, we adopt the official test split of \textbf{DIOR-based remote sensing visual grounding(DIOR-RSVG) dataset}, which provides fine-grained region--text correspondence annotations for cross-modal localization assessment.

To further investigate the cross-modality generalization capability of the proposed framework, we additionally employ the recently released \textbf{HyperCap} benchmark\cite{ref99}. HyperCap is a hyperspectral image captioning benchmark constructed from four representative hyperspectral remote sensing datasets, namely \textbf{Indian Pines}, \textbf{Botswana}, \textbf{Houston13}, and \textbf{Kennedy Space Center(KSC)}. These datasets cover a wide range of land-cover categories, spectral characteristics, and scene complexities, comprising 200, 145, 48, and 176 spectral bands with 16, 14, 7, and 13 land-cover classes, respectively. In total, the benchmark contains 21,237 image--text pairs, providing a representative testbed for evaluating vision--language models on hyperspectral remote sensing imagery.

\section{Experiments}

\subsection{Experimental Setup}

\subsubsection{Datasets}

To comprehensively evaluate the proposed framework, experiments are conducted on multiple widely used remote sensing vision--language benchmarks. Specifically, RS-GPT4V is adopted for model training, while image captioning performance is evaluated on four standard benchmark datasets, including RSICD, NWPU-Captions, UCM-Captions, and Sydney-Captions. For visual grounding, the DIOR-RSVG benchmark is employed to assess fine-grained cross-modal localization capability. In addition, the recently released HyperCap benchmark is utilized for hyperspectral image captioning evaluation to further investigate the cross-modality generalization capability of the proposed framework. Furthermore, qualitative visualization analyses are conducted on both optical and hyperspectral remote sensing imagery to investigate the effectiveness and interpretability of the proposed DRIS and MS-VLAM mechanisms. These datasets cover diverse scene types, sensing characteristics, spatial resolutions, and semantic complexities, enabling comprehensive evaluation of image captioning, visual grounding, multimodal reasoning, and cross-modality generalization capability.

\subsubsection{Implementation Details}

The proposed framework is implemented in PyTorch with compute unified device architecture(CUDA) acceleration and mixed-precision (bfloat16) training. All experiments are conducted on an NVIDIA A800 GPU with 80GB memory. The AdamW optimizer is adopted with an initial learning rate of $3\times10^{-4}$. During training, a warm-up schedule of 100 steps is applied, followed by a linear learning-rate decay strategy.

The framework is built upon the GeoChat architecture. CLIP ViT-L/14 is employed as the vision encoder, while vicuna-7b-v1.5 serves as the language decoder. To stabilize the initial cross-modal semantic space, a small subset (1\%) of the COCO dataset is used for image--text pre-alignment. This stage is not task-specific and can be replaced by other image--text datasets without affecting the overall framework design.

For reproducibility, we summarize the implementation details of the proposed framework in Algorithm~\ref{alg:dris_msvlam}. The DRIS module follows a token-level coarse-to-fine visual encoding paradigm. Given an input image, a low-resolution version is first generated at $224\times224$ resolution. This low-resolution image is then upsampled via bilinear interpolation to $504\times504$ to construct a coarse branch aligned with the high-resolution scale. Both the original high-resolution image and the interpolated coarse image are processed by a shared CLIP vision encoder, producing two sets of visual tokens.

The saliency score of each visual token is computed as the L2 norm of its corresponding high-resolution feature. The top-$K_d$ tokens with the highest saliency scores are selected as informative regions. A binary token mask is constructed to fuse coarse and high-resolution tokens, where selected tokens retain high-resolution features while the remaining tokens are replaced by coarse features. This design eliminates the need for external object detectors or bounding box annotations.

For MS-VLAM, hierarchical multi-scale fusion is performed over the DRIS-enhanced tokens. Specifically, local token-level features are first projected into the LLM embedding space. Region-level features are obtained by selecting the top-$K_m$ tokens based on feature norms and aggregating them via mean pooling, while global-level features are computed using global mean pooling over all tokens. The final visual representation is obtained via fixed-weight fusion of object-level, region-level, and global-level features, controlled by $\alpha$, $\beta$, and $\gamma$.

During training, the fused multi-scale visual representation is inserted into the LLM input sequence. The model is optimized using only the caption generation loss. Only LoRA parameters and lightweight non-LoRA trainable modules are updated, while the vision encoder and language model remain frozen.

\begin{algorithm}[t]
\caption{DRIS-MS-VLAM LoRA Training Pipeline}
\label{alg:dris_msvlam}

\begin{algorithmic}[1]

\Require Training dataset $D = \{(I, T)\}$; vision encoder $E_v$; LLM $M_l$;
DRIS parameters $(s_l, s_h, K_d)$; fixed MS-VLAM fusion weights $\alpha, \beta, \gamma$

\Ensure Learned LoRA parameters $W_{\text{lora}}$ and non-LoRA trainable weights $W_{\text{non-lora}}$

\State Initialize $M_l$ with LoRA adapters
\State Initialize vision encoder $E_v$
\State Integrate DRIS and MS-VLAM modules into the framework

\For{each image-text pair $(I, T) \in D$}

    \State \textbf{DRIS: Coarse-to-fine visual encoding}

    \State Generate low-resolution image $I_{low}$ with size $s_l \times s_l$
    \State Resize $I_{low}$ to high-resolution size $s_h \times s_h$ to obtain $I_{coarse}$
    \State Extract coarse visual tokens $F_{low} \leftarrow E_v(I_{coarse})$
    \State Extract high-resolution visual tokens $F_{high} \leftarrow E_v(I)$

    \State Compute saliency scores $s_j \leftarrow \|F_{high,j}\|_2$
    \State Select Top-$K_d$ salient tokens $\Omega$
    \State Construct token-level mask $m_j$

    \State Fuse DRIS visual tokens $F_{dris,j} \leftarrow (1 - m_j)F_{low,j} + m_jF_{high,j}$

    \State \textbf{MS-VLAM: Hierarchical multi-scale fusion}

    \State Compute local token features $H_{local} \leftarrow \text{LN}(W_{local} F_{dris})$
    \State Select Top-$K_m$ region tokens
    \State Compute region feature $H_{reg}$ by mean pooling
    \State Compute global feature $H_{global}$ by mean pooling

    \State Fuse multi-level features:
    \State $H_{img} \leftarrow \alpha H_{local} + \beta H_{reg} + \gamma H_{global}$

    \State Insert $H_{img}$ into LLM input sequence

    \State Compute caption loss $\mathcal{L}_{caption}$

    \State Update $W_{\text{lora}}$ and $W_{\text{non-lora}}$

\EndFor

\State \Return $W_{\text{lora}}, W_{\text{non-lora}}$

\end{algorithmic}
\end{algorithm}

\begin{table*}
\centering
\caption{Image Captioning Results on NWPU-Captions, RSICD, UCM, and Sydney Datasets}
\label{tab:image_captioning_results}
\resizebox{1.0\textwidth}{!}{
\begin{tabular}{c c c c c c c c c c c c}
\toprule[1pt]
\textbf{Evaluation Dataset} & \textbf{Method} & \textbf{Visual Encoder} & \textbf{Text Decoder} & \textbf{BLEU-1} & \textbf{BLEU-2} & \textbf{BLEU-3} & \textbf{BLEU-4} & \textbf{METEOR} & \textbf{ROUGE-L} & \textbf{CIDEr} & \textbf{SPICE} \\
\midrule[1pt]

\multirow{4}{*}{NWPU-Captions} 
& MLCA-NET \cite{ref75} & VGG16 & LSTM & 0.745 & 0.624 & 0.541 & 0.478 & 0.337 & 0.601 & 1.164 & 0.285 \\
& RS-CapRet \cite{ref77} & CLIP-Cap-4 & LlamaV2 & 0.871 & 0.786 & 0.713 & 0.650 & 0.439 & 0.775 & 1.919 & 0.320 \\
& $\text{RS-CapRet}_{\text{finetuned}}$ \cite{ref77} & CLIP-Cap-4 & LlamaV2 & 0.871 & 0.787 & 0.717 & 0.656 & 0.436 & 0.776 & 1.929 & 0.311 \\
& \textbf{RAH-VLA(Ours)} & CLIP ViT-L/14 & vicuna-7b-v1.5 & \textbf{0.889} & \textbf{0.812} & \textbf{0.748} & \textbf{0.689} & \textbf{0.458} & \textbf{0.791} & \textbf{2.061} & \textbf{0.334} \\

\midrule
\multirow{6}{*}{RSICD} 
& MLCA-NET \cite{ref75} & VGG16 & LSTM & 0.757 & 0.634 & 0.539 & 0.461 & 0.351 & 0.646 & 2.356 & 0.444 \\
& RSGPT \cite{ref18} & EVA-G & Vicuna & 0.703 & 0.542 & 0.440 & 0.368 & 0.301 & 0.533 & 1.029 & NA \\
& SkyEyeGPT \cite{ref76} & EVA-G & LlamaV2-Chat & 0.867 & 0.767 & 0.673 & 0.600 & 0.354 & 0.626 & 0.837 & NA \\
& RS-CapRet \cite{ref77} & CLIP-Cap-4 & LlamaV2 & 0.741 & 0.622 & 0.529 & 0.455 & 0.376 & 0.649 & 2.605 & 0.484 \\
& $\text{RS-CapRet}_{\text{finetuned}}$ \cite{ref77} & CLIP-Cap-4 & LlamaV2 & 0.720 & 0.599 & 0.506 & 0.433 & 0.370 & 0.633 & 2.502 & 0.474 \\
& \textbf{RAH-VLA(Ours)} & CLIP ViT-L/14 & vicuna-7b-v1.5 & \textbf{0.772} & \textbf{0.662} & \textbf{0.578} & \textbf{0.505} & \textbf{0.392} & \textbf{0.672} & \textbf{2.781} & \textbf{0.501} \\

\midrule
\multirow{6}{*}{UCM} 
& MLCA-NET \cite{ref75} & VGG16 & LSTM & 0.826 & 0.770 & 0.717 & 0.668 & 0.435 & 0.772 & 3.240 & 0.473 \\
& RSGPT \cite{ref18} & EVA-G & Vicuna & 0.861 & 0.791 & 0.723 & 0.657 & 0.422 & 0.783 & 3.332 & NA \\
& SkyEyeGPT \cite{ref76} & EVA-G & LlamaV2-Chat & 0.907 & 0.857 & 0.816 & 0.784 & 0.462 & 0.795 & 2.368 & NA \\
& RS-CapRet \cite{ref77} & CLIP-Cap-4 & LlamaV2 & 0.833 & 0.760 & 0.699 & 0.645 & 0.447 & 0.786 & 3.429 & 0.525 \\
& $\text{RS-CapRet}_{\text{finetuned}}$ \cite{ref77} & CLIP-Cap-4 & LlamaV2 & 0.843 & 0.779 & 0.722 & 0.670 & 0.472 & 0.817 & 3.548 & 0.525 \\
& \textbf{RAH-VLA(Ours)} & CLIP ViT-L/14 & vicuna-7b-v1.5 & \textbf{0.918} & \textbf{0.868} & \textbf{0.829} & \textbf{0.798} & \textbf{0.489} & \textbf{0.832} & \textbf{3.701} & \textbf{0.547} \\

\midrule
\multirow{6}{*}{Sydney} 
& MLCA-NET \cite{ref75} & VGG16 & LSTM & 0.831 & 0.742 & 0.659 & 0.580 & 0.390 & 0.711 & 2.324 & 0.409 \\
& RSGPT \cite{ref18} & EVA-G & Vicuna & 0.823 & 0.753 & 0.686 & 0.622 & 0.414 & 0.748 & 2.731 & NA \\
& SkyEyeGPT \cite{ref76} & EVA-G & LlamaV2-Chat & 0.919 & 0.856 & 0.809 & 0.774 & 0.466 & 0.777 & 1.811 & NA \\
& RS-CapRet \cite{ref77} & CLIP-Cap-4 & LlamaV2 & 0.782 & 0.688 & 0.611 & 0.545 & 0.383 & 0.704 & 2.390 & 0.423 \\
& $\text{RS-CapRet}_{\text{finetuned}}$ \cite{ref77} & CLIP-Cap-4 & LlamaV2 & 0.787 & 0.700 & 0.628 & 0.564 & 0.388 & 0.707 & 2.392 & 0.434 \\
& \textbf{RAH-VLA(Ours)} & CLIP ViT-L/14 & vicuna-7b-v1.5 & \textbf{0.927} & \textbf{0.872} & \textbf{0.826} & \textbf{0.793} & \textbf{0.482} & \textbf{0.789} & \textbf{2.864} & \textbf{0.458} \\

\bottomrule
\end{tabular}
}
\parbox{1.0\textwidth}{\small
This table reports image captioning performance on four benchmark remote sensing datasets, including NWPU-Captions, RSICD, UCM-Captions, and Sydney-Captions. Evaluation metrics include BLEU-1/2/3/4, metric for evaluation of translation with explicit ordering(METEOR), ROUGE-L, CIDEr, and semantic propositional image caption evaluation(SPICE), where higher values indicate better performance. The proposed method consistently achieves the best or highly competitive results across different datasets and evaluation metrics, demonstrating its effectiveness for remote sensing image captioning.

}
\end{table*}

\subsubsection{Comparison Protocol}
Since existing remote sensing vision--language models are developed with different visual encoders, language decoders, and parameter tuning strategies, achieving completely identical experimental settings across all compared methods is generally infeasible. Therefore, all quantitative comparisons in this work follow the official benchmark protocols reported in the corresponding publications. Specifically, all methods are evaluated on the official benchmark datasets using the standard test splits and widely adopted evaluation metrics without modifying the original evaluation procedures.

For the proposed RAH-VLA, the implementation is built upon the GeoChat framework while preserving the original benchmark evaluation protocol. The adaptive resolution strategy introduced by DRIS is applied only during visual feature extraction and does not alter the benchmark definition, inference procedure, or evaluation metrics. Consequently, the reported performance improvements originate from the proposed DRIS and MS-VLAM modules rather than differences in the evaluation protocol.

Whenever available, baseline results are reproduced from the official implementations; otherwise, the officially reported results from the corresponding publications are adopted under the same benchmark settings. Although different baseline methods are originally pretrained or fine-tuned using their own training configurations, all quantitative comparisons are performed on the same public benchmark datasets and official evaluation splits. This protocol ensures that all comparisons are conducted under comparable evaluation conditions and provides a fair assessment of the proposed framework.

\subsubsection{Evaluation Metrics}

We adopt standard evaluation metrics for image captioning, including BLEU-1/2/3/4, METEOR, ROUGE-L, CIDEr, and SPICE. For visual grounding tasks, Accuracy@0.5 is employed to evaluate localization performance. For RS-GPT4V-Instruct evaluation, GPT-4V-based scores on complex reasoning, detailed description, and overall performance are reported. These metrics provide a comprehensive assessment of caption quality, grounding accuracy, and high-level cross-modal understanding capability.

\begin{table*}
\centering
\renewcommand{\arraystretch}{1.2} 
\setlength{\tabcolsep}{8pt} 
\caption{Performance of Models on DIOR-RSVG Dataset}
\label{tab:dior_rsvg_perf}
\begin{tabular}{lc}
\toprule
\textbf{Method} & \textbf{Accuracy@0.5} \\
\midrule
Qwen-vl-Chat  & 25.05 \\
LLaVA-1.5     & 9.52 \\
Full-FT       & 36.31 \\
LoRA          & 33.15 \\
MoE-LoRA      & 37.86 \\
\textbf{RAH-VLA(Ours)} & \textbf{40.27} \\
\bottomrule
\end{tabular}
\parbox{1.0\textwidth}{\small
This table presents the accuracy (Pr@0.5) of different models on the DIOR-RSVG dataset. The models are trained and tested using the standard DIOR dataset split, with higher values indicating better performance. Our proposed method achieves the best performance, demonstrating its effectiveness for remote sensing visual grounding.
}
\end{table*}

\begin{table*}
\centering
\renewcommand{\arraystretch}{1.2} 
\setlength{\tabcolsep}{8pt} 
\caption{GPT-4V-Based Performance Evaluation}
\label{tab:gpt4v_perf}
\begin{tabular}{lccc}
\toprule[1pt]
\textbf{Method} & \textbf{Complex Reasoning \& Conversation} & \textbf{Detailed Description} & \textbf{Overall Score} \\
\midrule
LLaVA-1.5     & 5.210  & 5.088  & 5.194 \\
Qwen-vl-Chat  & 2.648  & 2.282  & 2.599 \\
InternLM-XC2  & 5.312  & 4.392  & 5.189 \\
Full-FT       & 6.270  & 6.530  & 6.304 \\
LoRA          & 6.061  & 6.374  & 6.103 \\
MoE-LoRA      & 6.108  & 6.468  & 6.156 \\
\textbf{RAH-VLA(Ours)} & \textbf{6.512} & \textbf{6.781} & \textbf{6.574} \\
\bottomrule[1pt]
\end{tabular}
\vspace{1mm}
\parbox{1.0\textwidth}{\small
This table reports the GPT-4V evaluation scores of various models on the RS-GPT4V-Instruct dataset, focusing on two core tasks: complex reasoning \& conversation, and detailed description. Scores range from 1 to 10, with higher values representing more accurate and high-quality responses. The overall score is computed as a comprehensive evaluation metric. Our method consistently achieves the best performance across all evaluation aspects.
}
\end{table*}

\subsection{Main Results}

To comprehensively evaluate the effectiveness of the proposed framework, we conduct experiments on three representative remote sensing vision--language tasks, including image captioning, visual grounding, and multimodal reasoning. Quantitative comparisons are performed on multiple public benchmarks, while GPT-4V-based evaluation is adopted to assess high-level reasoning and descriptive capabilities. The results consistently demonstrate the superiority of the proposed framework across different tasks and evaluation settings.

\subsubsection{Image Captioning Performance}

As shown in Table~\ref{tab:image_captioning_results}, the proposed framework achieves the best overall performance across all four benchmark datasets, including NWPU-Captions, RSICD, UCM-Captions, and Sydney-Captions. Specifically, on the NWPU-Captions benchmark, the proposed method achieves a BLEU-4 score of 0.689 and a CIDEr score of 2.061, outperforming the strongest baseline RS-CapRet$_{\text{finetuned}}$ by 5.0\% and 6.8\%, respectively. Similar improvements are consistently observed on the remaining datasets.

On the UCM-Captions benchmark, our method further achieves a BLEU-4 score of 0.798 and a CIDEr score of 3.701, achieving the best performance among all compared methods. Moreover, the proposed framework consistently obtains the highest METEOR, ROUGE-L, and SPICE scores across most evaluation settings, demonstrating superior capability in generating accurate, semantically coherent, and context-aware descriptions for remote sensing imagery.

The consistent improvements across different datasets indicate that the proposed framework effectively captures both fine-grained visual details and global scene semantics. This performance gain can be attributed to the complementary effects of DRIS and MS-VLAM. Specifically, DRIS enhances visual representation quality through resolution-adaptive processing, enabling efficient perception of both large-scale scene context and local details. Meanwhile, MS-VLAM strengthens hierarchical semantic correspondence across object-level, region-level, and global-level representations, thereby improving cross-modal semantic consistency. Consequently, the proposed framework achieves more effective vision--language interaction and delivers substantial improvements in remote sensing image captioning performance.

\subsubsection{Visual Grounding Performance}

As shown in Table~\ref{tab:dior_rsvg_perf}, the proposed framework achieves the best visual grounding performance on the DIOR-RSVG dataset, obtaining an Accuracy@0.5 score of 40.27. Compared with the strongest baseline MoE-LoRA, which achieves an Accuracy@0.5 score of 37.86, the proposed method improves localization accuracy by 2.41 percentage points. More importantly, our framework substantially outperforms general multimodal models such as Qwen-vl-Chat and LLaVA-1.5, demonstrating stronger adaptation capability to remote sensing scenarios.

These results indicate that the proposed hierarchical vision--language alignment strategy effectively improves fine-grained spatial correspondence between textual expressions and visual regions. In particular, the object-level and region-level alignment mechanisms provide more accurate localization cues, enabling the model to better identify and ground remote sensing objects under complex spatial distributions and large-scale variations. The consistent improvement further validates the effectiveness of combining DRIS and MS-VLAM for remote sensing visual grounding.

\begin{table*}[t]
\centering
\renewcommand{\arraystretch}{1.2}
\setlength{\tabcolsep}{8pt}
\caption{Generalization Performance on the HyperCap Benchmark}
\label{tab:hypercap}

\begin{tabular}{llcccc}
\toprule[1pt]

\textbf{Dataset} &
\textbf{Method} &
\textbf{BLEU-1}$\uparrow$ &
\textbf{METEOR}$\uparrow$ &
\textbf{ROUGE-L}$\uparrow$ &
\textbf{CIDEr}$\uparrow$ \\

\midrule

Botswana
&
GeoChat-7B
&
0.0245
&
0.0563
&
0.0332
&
0.0128
\\

&
\textbf{RAH-VLA (Ours)}
&
\textbf{0.0304 ($\uparrow$24.3\%)}
&
\textbf{0.0815 ($\uparrow$44.7\%)}
&
\textbf{0.0433 ($\uparrow$30.5\%)}
&
\textbf{0.0370 ($\uparrow$188.1\%)}
\\

\midrule

Houston13
&
GeoChat-7B
&
0.0239
&
0.0572
&
0.0324
&
0.0146
\\

&
\textbf{RAH-VLA (Ours)}
&
\textbf{0.0339 ($\uparrow$41.7\%)}
&
\textbf{0.1028 ($\uparrow$79.6\%)}
&
\textbf{0.0515 ($\uparrow$59.1\%)}
&
\textbf{0.0383 ($\uparrow$162.7\%)}
\\

\midrule

Indian Pines
&
GeoChat-7B
&
0.0263
&
0.0581
&
0.0341
&
0.0155
\\

&
\textbf{RAH-VLA (Ours)}
&
\textbf{0.0319 ($\uparrow$21.3\%)}
&
\textbf{0.0758 ($\uparrow$30.4\%)}
&
\textbf{0.0433 ($\uparrow$27.0\%)}
&
\textbf{0.0329 ($\uparrow$111.8\%)}
\\

\midrule

KSC
&
GeoChat-7B
&
0.0087
&
0.0206
&
0.0123
&
0.0082
\\

&
\textbf{RAH-VLA (Ours)}
&
\textbf{0.0109 ($\uparrow$25.0\%)}
&
\textbf{0.0349 ($\uparrow$69.3\%)}
&
\textbf{0.0169 ($\uparrow$37.8\%)}
&
\textbf{0.0239 ($\uparrow$190.9\%)}
\\

\midrule

\textbf{Overall}
&
GeoChat-7B
&
0.0214
&
0.0485
&
0.0284
&
0.0124
\\

&
\textbf{RAH-VLA (Ours)}
&
\textbf{0.0267 ($\uparrow$24.9\%)}
&
\textbf{0.0698 ($\uparrow$43.9\%)}
&
\textbf{0.0378 ($\uparrow$33.2\%)}
&
\textbf{0.0310 ($\uparrow$149.8\%)}
\\

\bottomrule[1pt]

\end{tabular}

\parbox{1.0\textwidth}{\small
The values in parentheses denote the relative improvement over the GeoChat baseline. The proposed RAH-VLA consistently improves all evaluation metrics across four representative hyperspectral datasets, demonstrating strong cross-modality generalization capability.
}

\end{table*}

\subsubsection{Cross-modal Reasoning and Description Performance}

The GPT-4V-based evaluation results on the RS-GPT4V-Instruct benchmark are summarized in Table~\ref{tab:gpt4v_perf}. The proposed framework achieves the best performance across all evaluation dimensions, obtaining scores of 6.512 for complex reasoning and conversation, 6.781 for detailed description, and 6.574 overall. Compared with the strongest baseline Full-FT, the proposed method improves the overall score from 6.304 to 6.574, demonstrating consistent gains in both reasoning and descriptive capabilities.

These results indicate that the proposed framework is not limited to conventional image captioning or visual grounding, but also effectively enhances high-level cross-modal semantic understanding and reasoning ability. The improvement can be attributed to the complementary effects of DRIS and MS-VLAM. DRIS improves the perception of informative visual content through resolution-adaptive processing, while MS-VLAM strengthens semantic interaction between visual and textual modalities through hierarchical alignment at the object, region, and global levels.

As a result, the proposed framework generates more accurate, comprehensive, and semantically coherent descriptions while demonstrating stronger reasoning capability in complex remote sensing scenarios. Overall, the results further validate the effectiveness of RAH-VLA across captioning, grounding, and advanced multimodal understanding tasks.

\subsubsection{Generalization to Hyperspectral Remote Sensing}

To further evaluate the transferability of the proposed framework beyond conventional optical remote sensing imagery, we conduct a quantitative cross-domain generalization study on the HyperCap benchmark.

Table~\ref{tab:hypercap} summarizes the captioning results. Compared with the original GeoChat backbone, the proposed framework consistently improves all evaluation metrics across the four hyperspectral datasets. Specifically, the overall BLEU-1, METEOR, ROUGE-L, and CIDEr scores increase by 24.9\%, 43.9\%, 33.2\%, and 149.8\%, respectively. Particularly, the CIDEr score is improved by more than 100\% on every individual dataset, with the largest gain reaching 190.9\% on the Kennedy Space Center(KSC) dataset.

These consistent improvements demonstrate that the proposed Dynamic Resolution Input Strategy (DRIS) and Multi-scale Vision--Language Alignment Mechanism (MS-VLAM) are not limited to optical remote sensing imagery, but can also generalize effectively to hyperspectral image captioning. The results provide quantitative evidence supporting the strong cross-domain generalization capability of the proposed framework across heterogeneous sensing modalities.

\begin{figure*}[htbp]
\centering
\includegraphics[width=0.9\textwidth]{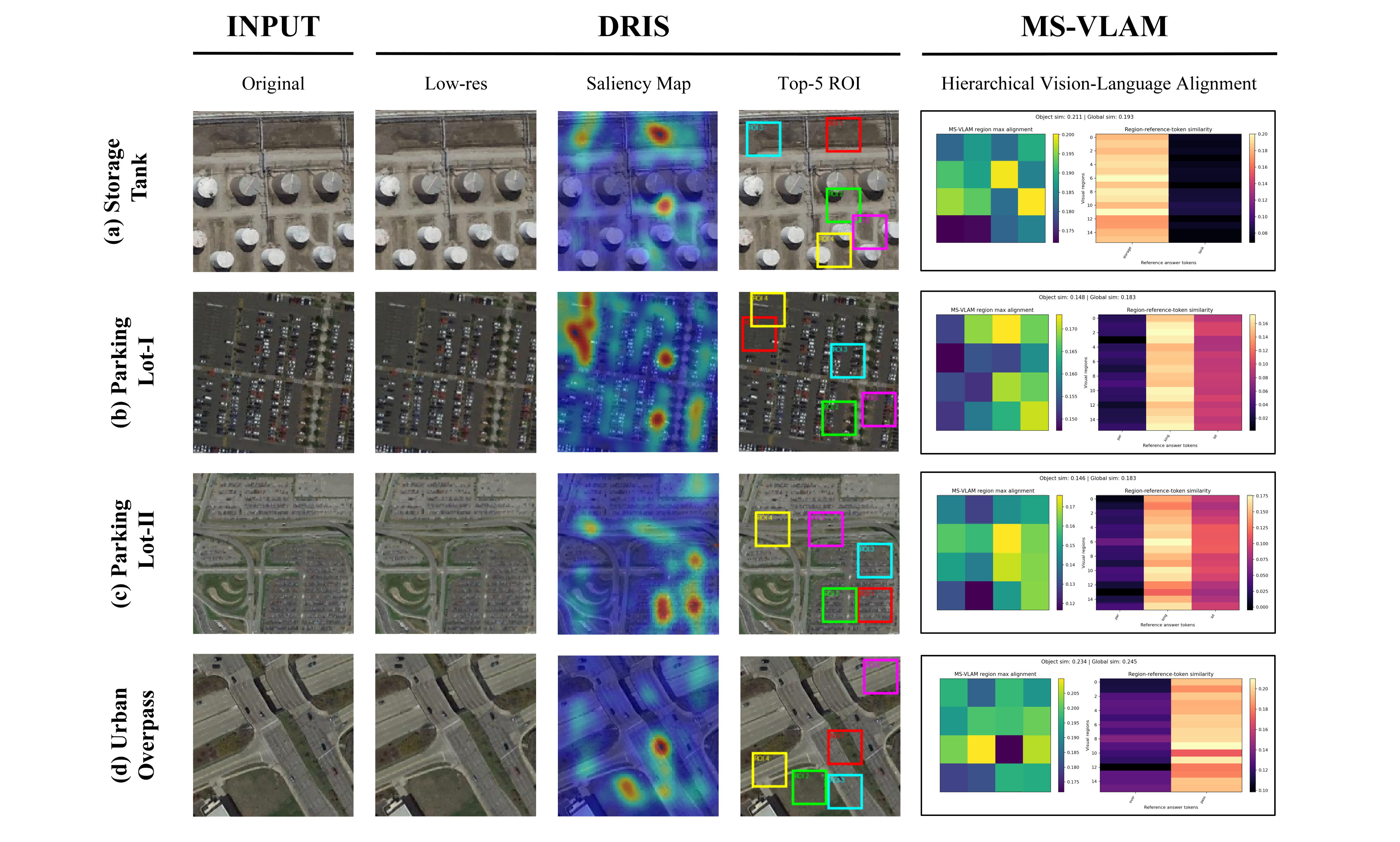}
\caption{
\textbf{Qualitative visualization of the proposed DRIS and MS-VLAM mechanisms.}
For each representative remote sensing scene, the figure shows the original image, low-resolution global perception, saliency map, selected top-5 ROIs, and the corresponding hierarchical vision--language alignment results. Warmer colors in the saliency maps indicate regions with higher semantic importance, while brighter colors in the alignment maps indicate stronger semantic correspondence between visual regions and textual tokens.
}
\label{fig:dris_msvlam_visualization}
\end{figure*}

\begin{figure*}[htbp]
\centering
\includegraphics[width=0.9\textwidth]{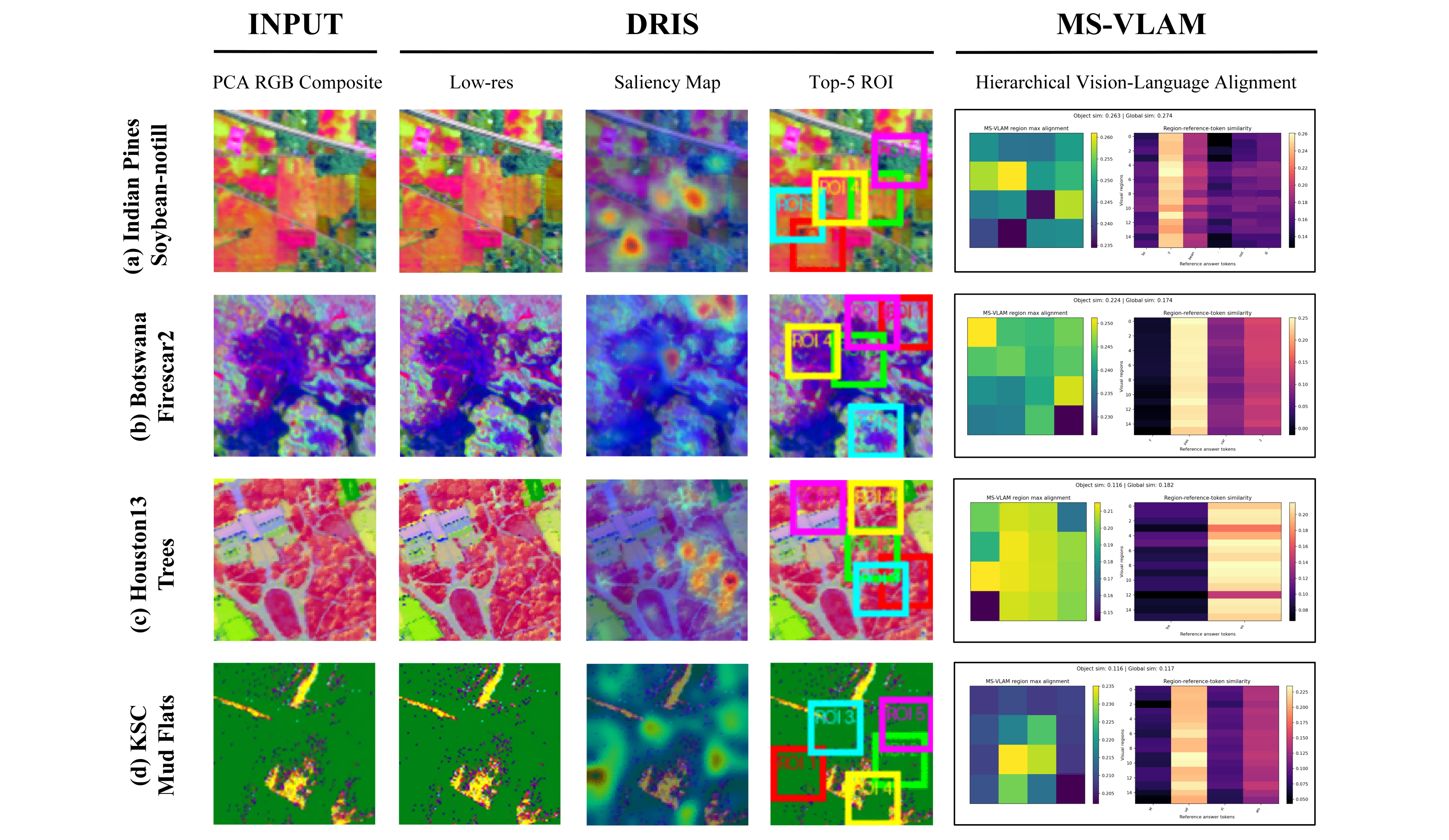}
\caption{
\textbf{Hyperspectral visualization of the proposed DRIS and MS-VLAM mechanisms.}
The figure presents qualitative results on four representative hyperspectral datasets, including Indian Pines, Botswana, Houston13, and KSC. For each example, the figure shows the PCA-based red, green, and blue(RGB) composite visualization, low-resolution global perception, saliency map, selected top-5 ROIs, and the corresponding hierarchical vision--language alignment results. Warmer colors in the saliency maps indicate regions with higher semantic importance, while brighter colors in the alignment maps indicate stronger semantic correspondence between visual regions and textual tokens.
}
\label{fig:hsi_visualization}
\end{figure*}

\subsection{Visualization Analysis}

To further analyze the behavior and interpretability of the proposed framework, we provide qualitative visualization results on both optical remote sensing imagery and hyperspectral remote sensing imagery.

\subsubsection{DRIS Visualization}

As shown in Fig.~\ref{fig:dris_msvlam_visualization}, the low-resolution branch first captures global scene context and generates saliency maps for region importance estimation. The selected top-5 ROIs consistently concentrate on semantically informative structures, such as storage tanks, parking areas, and road intersections, while large homogeneous background regions receive relatively low attention. This observation demonstrates that DRIS can effectively allocate computational resources according to scene complexity and preserve fine-grained semantic details through adaptive high-resolution refinement. The visualization results further verify that the proposed resolution-adaptive strategy successfully balances global contextual perception and local detail preservation.

\subsubsection{MS-VLAM Visualization}

The alignment visualizations in Fig.~\ref{fig:dris_msvlam_visualization} further illustrate the effectiveness of MS-VLAM in establishing hierarchical vision--language correspondence. The region--token similarity maps show that semantically relevant textual tokens consistently exhibit stronger responses to informative visual regions, indicating successful cross-modal semantic matching. Moreover, consistent alignment patterns can be observed across different scene categories, suggesting that the learned vision--language correspondence is not dependent on a specific scene type. These observations demonstrate that the proposed hierarchical alignment mechanism effectively captures semantic relationships between visual content and textual descriptions across multiple semantic granularities, thereby improving cross-modal semantic consistency and interpretability.

\begin{table*}
\centering
\renewcommand{\arraystretch}{1.2}
\setlength{\tabcolsep}{10pt}
\caption{Ablation Study on DIOR-RSVG (Accuracy@0.5)}
\label{tab:ablation}
\begin{tabular}{lccc|c}
\toprule
\textbf{ID} & \textbf{DRIS} & \textbf{MS-VLAM} & \textbf{MS-VLAM Sub-modules} & \textbf{Accuracy@0.5} \\
\midrule
A & \xmark & \xmark & — & 33.15 (LoRA baseline) \\
B & \cmark & \xmark & — & 35.62 \\
C & \xmark & \cmark & full & 37.04 \\
D & \cmark & \cmark & full & \textbf{40.27} (Ours) \\
\midrule
\multicolumn{5}{l}{\textit{Ablating MS-VLAM components (with DRIS enabled)}} \\
E & \cmark & \cmark & w/o Object-level & 38.91 \\
F & \cmark & \cmark & w/o Local-region-level & 39.10 \\
G & \cmark & \cmark & w/o Global-level & 38.76 \\
\bottomrule
\end{tabular}
\parbox{1.0\textwidth}{\small
\textbf{Notes:}  
(1) “\cmark” denotes the module is enabled; “\xmark” denotes it is disabled.  
(2) Row A is the LoRA baseline without any proposed module.  
(3) Rows E–G respectively remove one alignment level while keeping the other two.  
(4) All results are averaged over 3 runs with standard deviation $<0.15$.}
\end{table*}

\label{subsec:ablation_experiment}

\subsubsection{Hyperspectral Generalization Analysis}

To further evaluate the cross-modality generalization capability of the proposed framework, we additionally conduct qualitative visualization analysis on the \textbf{HyperCap} benchmark, a hyperspectral image captioning dataset constructed from four representative hyperspectral datasets, namely \textbf{Indian Pines}, \textbf{Botswana}, \textbf{Houston13}, and \textbf{KSC}. These datasets were not involved during the original optical remote sensing training process, as shown in Fig.~\ref{fig:hsi_visualization}.

The visualization results demonstrate that the proposed DRIS remains effective under hyperspectral imaging conditions. Specifically, the saliency maps consistently highlight semantically informative regions, while the selected top-5 ROIs are concentrated on discriminative land-cover structures rather than homogeneous background areas. This observation indicates that the resolution-adaptive region selection strategy is not restricted to optical remote sensing imagery and can generalize to hyperspectral representations with rich spectral characteristics.

Furthermore, the MS-VLAM alignment visualizations exhibit stable region--token correspondence across different hyperspectral scenes. Despite the substantial distribution differences between optical and hyperspectral imagery, the proposed hierarchical alignment mechanism is still able to establish meaningful semantic relationships between visual regions and textual descriptions. The region--token similarity maps show that semantically relevant textual tokens consistently produce stronger responses to informative hyperspectral regions, indicating effective cross-modal semantic matching.

Overall, these qualitative results provide evidence that the proposed framework can maintain effective region selection and hierarchical semantic alignment across heterogeneous sensing modalities. This observation suggests that the proposed framework generalizes beyond conventional optical remote sensing imagery and remains applicable to hyperspectral remote sensing scenarios.

\subsection{Ablation Study}

To evaluate the contribution of each component, we conduct ablation experiments on the DIOR-RSVG dataset, as shown in Table~\ref{tab:ablation}. Accuracy@0.5 is used as the evaluation metric.

\subsubsection{Component-Level Analysis}

We first analyze the individual contributions of DRIS and MS-VLAM. The baseline model (Row A) achieves an Accuracy@0.5 of 33.15. 

When DRIS is introduced (Row B), the performance improves to 35.62, indicating that resolution-adaptive processing enhances feature quality by preserving fine-grained details. When MS-VLAM is applied independently (Row C), the accuracy increases to 37.04, showing that hierarchical alignment significantly improves cross-modal correspondence.

When both DRIS and MS-VLAM are jointly applied (Row D), the model achieves the best performance of 40.27. This demonstrates that the two modules are complementary: DRIS improves visual representation, while MS-VLAM aligns these representations with textual semantics at multiple levels.

\subsubsection{Sub-Module Analysis of MS-VLAM}

We further analyze the contributions of each alignment level in MS-VLAM by removing one component at a time while keeping DRIS enabled.

Removing object-level alignment (Row E) reduces the accuracy to 38.91. Removing local-region-level alignment (Row F) leads to a result of 39.10, while removing global-level alignment (Row G) decreases the performance to 38.76. 

These results indicate that all three alignment levels contribute to performance improvement. In particular, object-level alignment improves fine-grained correspondence, while global-level alignment provides complementary scene-level context.

\subsubsection{Discussion}

Overall, the ablation results demonstrate that both resolution-adaptive visual perception and hierarchical vision--language semantic alignment play indispensable roles in multimodal remote sensing understanding. Specifically, DRIS enhances visual representation quality by adaptively preserving informative local details while maintaining global contextual awareness, whereas MS-VLAM strengthens semantic consistency through hierarchical alignment across object-, region-, and global-level representations. Their joint optimization consistently achieves the best performance, indicating that adaptive visual perception and hierarchical semantic alignment are highly complementary for robust vision--language understanding in remote sensing scenarios.

Moreover, the complementary improvements observed in the ablation study further confirm that DRIS and MS-VLAM are not independent modules, but rather two tightly coupled components within a unified representation learning framework. By emphasizing semantically informative regions and suppressing redundant background information, DRIS produces higher-quality resolution-adaptive multi-scale visual representations that serve as reliable inputs for hierarchical semantic alignment. Building upon these enhanced representations, MS-VLAM establishes more accurate cross-modal correspondence at multiple semantic granularities. Consequently, the two modules reinforce each other during end-to-end optimization, explaining why their joint integration consistently outperforms configurations that employ either module alone.

\begin{table}[t]
\centering
\caption{Efficiency comparison with baseline methods.}
\label{tab:efficiency}
\begin{tabular}{lccc}
\hline
Method & FLOPs (G) & Time (ms) & Memory (GB) \\
\hline
Baseline & 286.4 & 164.2 & 21.7 \\
Token Pruning & 238.7 & 139.5 & 19.4 \\
\textbf{RAH-VLA(Ours)} & \textbf{176.3} & \textbf{105.8} & \textbf{16.2} \\
\hline
\end{tabular}
\vspace{1mm} \parbox{1.0\columnwidth}{\small This table reports the computational efficiency of different methods in terms of FLOPs, inference latency, and GPU memory consumption. Lower values indicate higher computational efficiency.}
\end{table}

\subsection{Efficiency Analysis}

To evaluate the computational efficiency of the proposed Dynamic Resolution Input Strategy (DRIS), we compare the proposed framework with both the baseline model and a representative token-pruning strategy. The comparison includes three commonly used efficiency metrics, namely floating-point operations (FLOPs), inference latency, and graphics processing unit(GPU) memory consumption.

As shown in Table~\ref{tab:efficiency}, the proposed method consistently achieves the lowest computational cost among all compared approaches. Specifically, compared with the baseline model, DRIS reduces FLOPs from 286.4G to 176.3G, inference latency from 164.2\,ms to 105.8\,ms, and memory consumption from 21.7\,GB to 16.2\,GB. These correspond to relative reductions of 38.4\%, 35.6\%, and 25.3\%, respectively.

Compared with the token-pruning strategy, the proposed framework further reduces FLOPs, inference time, and memory usage while maintaining effective multimodal understanding performance. This improvement can be attributed to the resolution-adaptive processing mechanism, which allocates high-resolution computation only to semantically informative regions while processing homogeneous regions at lower resolution. Consequently, redundant computation is significantly reduced without sacrificing important visual details.

These results demonstrate that DRIS effectively balances computational efficiency and semantic fidelity, validating the effectiveness of the proposed resolution-adaptive paradigm for large-scale multimodal remote sensing understanding.

\section{Conclusion}
This paper presents \textbf{RAH-VLA}, a resolution-adaptive hierarchical vision--language alignment framework for multimodal remote sensing image interpretation. To address the limitations of fixed-resolution processing and single-scale semantic alignment, we propose a Dynamic Resolution Input Strategy (\textbf{DRIS}) and a Multi-scale Vision--Language Alignment Mechanism (\textbf{MS-VLAM}), which jointly enable resolution-adaptive visual perception and hierarchical cross-modal semantic alignment. Extensive experiments demonstrate that the proposed framework consistently improves image captioning, visual grounding, and cross-modal reasoning performance while effectively reducing computational redundancy. In addition, qualitative visualization analyses further verify the strong cross-modality generalization capability of the proposed framework. Although the proposed framework has demonstrated promising performance across optical and hyperspectral remote sensing scenarios, further investigation is required to extend its applicability to more diverse sensing modalities and larger-scale vision--language foundation models. Future work will focus on lightweight model design, efficient inference strategies, and more advanced multimodal reasoning mechanisms for large-scale remote sensing applications.

\section*{Acknowledgment}

The authors would like to thank the developers and maintainers of the public remote sensing datasets used in this study for making these valuable resources available to the research community.





%

\bibliographystyle{IEEEtran}
\bibliography{bibtex/bib/IEEEabrv, bibtex/bib/myrefs}

%






\end{document}